\documentclass{nle}

\usepackage{url}
\usepackage{graphicx}
\usepackage{wrapfig}
\usepackage{subfigure}
\usepackage{color}
\usepackage{amsmath}
\usepackage{amssymb}
\usepackage{amsfonts,eucal,amsbsy,amsthm,amsopn}
\usepackage{sidecap}
\usepackage{multirow}
\usepackage{algorithm}
\usepackage{algorithmic}
\usepackage{pdflscape}
\usepackage{afterpage}
\usepackage{longtable}

\usepackage{booktabs}
\newcommand{\tabitem}{~~\llap{\textbullet}~~}

\makeatletter
\newcommand{\longdash}[1][2em]{%
  \makebox[#1]{$\m@th\smash-\mkern-7mu\cleaders\hbox{$\mkern-2mu\smash-\mkern-2mu$}\hfill\mkern-7mu\smash-$}}
\makeatother
\newcommand{\omitskip}{\kern-\arraycolsep}

\newcommand{\hilighty}[1]{\colorbox{yellow}{#1}}
\newcommand{\hilightg}[1]{\colorbox{green}{#1}}
\newcommand{\hilightr}[1]{\colorbox{red}{#1}}
\newcommand{\hilightb}[1]{\colorbox{cyan}{#1}}
\newcommand{\hilightm}[1]{\colorbox{magenta}{#1}}

\let\cite\cite

\title[A Novel ILP Framework for Summarization]
      {A Novel ILP Framework for Summarizing Content with High Lexical Variety\thanks{This research is supported by an internal grant from the Learning Research and Development Center at the University of Pittsburgh as well as by an Andrew Mellon Predoctoral Fellowship to the first author. We are grateful to Logan Lebanoff for helping with the experiments. We also thank Muhsin Menekse, the CourseMIRROR team, and Wenting Xiong for providing or helping to collect some of our datasets. We thank Jingtao Wang, Fan Zhang, Huy Nguyen and Zahra Rahimi for valuable suggestions about the proposed summarization algorithm.}}

\author[Wencan Luo, Fei Liu, Zitao Liu, and Diane Litman]
{
W\ls E\ls N\ls C\ls A\ls N\ns L\ls U\ls O$^1$,\ns F\ls E\ls I\ns L\ls I\ls U$^2$,\ns Z\ls I\ls T\ls A\ls O\ns L\ls I\ls U$^3$\ls and\ns D\ls I\ls A\ls N\ls E\ns L\ls I\ls T\ls M\ls A\ls N$^1$\\
$^1$University of Pittsburgh, Pittsburgh, PA 15260\\
$^2$University of Central Florida, Orlando, FL 32816\\
$^3$Pinterest, Inc., San Francisco, CA 94103\\
e-mail: {\tt \{wencan, litman\}@cs.pitt.edu  \quad feiliu@cs.ucf.edu \quad zitaoliu@pinterest.com}
}

\begin{document}

\label{firstpage}
\maketitle

\begin{abstract}

Summarizing content contributed by individuals can be challenging, because people make different lexical choices even when describing the same events.
However, there remains a significant need to summarize such content. 
Examples include the student responses to post-class reflective questions, product reviews, and news articles published by different news agencies related to the same events.
High lexical diversity of these documents hinders the system's ability to effectively identify salient content and reduce summary redundancy. 
In this paper, we overcome this issue by introducing an integer linear programming-based summarization framework.
It incorporates a low-rank approximation to the sentence-word co-occurrence matrix to intrinsically group semantically-similar lexical items.
We conduct extensive experiments on datasets of student responses, product reviews, and news documents. 
Our approach compares favorably to a number of extractive baselines as well as a neural abstractive summarization system. 
The paper finally sheds light on when and why the proposed framework is effective at summarizing content with high lexical variety.


\end{abstract}

\section{Introduction}
\label{sec:intro}

Summarization is a promising technique for reducing information overload.
It aims at converting long text documents to short, concise summaries conveying the essential content of the source documents~\cite{Nenkova:2011}. 
Extractive methods focus on selecting important sentences from the source and concatenating them to form a summary, whereas abstractive methods can involve a number of high-level text operations such as word reordering, paraphrasing, and generalization~\cite{Jing:1999}.
To date, summarization has been successfully exploited for a number of text domains, including news articles~\cite{Barzilay:1999,Dang:2008,Durrett:2016:ACL,grusky2018newsroom}, product reviews~\cite{Gerani:2014}, online forum threads~\cite{Tarnpradab:2017}, meeting transcripts~\cite{Liu:2013:IEEETrans}, scientific articles~\cite{Teufel:2002,Qazvinian:2013}, student course responses~\cite{Luo:2015:EMNLP,Luo:2016:NAACL}, and many others.

Summarizing content contributed by multiple authors is particularly challenging. 
This is partly because people tend to use different expressions to convey the same semantic meaning. 
In a recent study of summarizing student responses to post-class reflective questions, Luo et al.,~\shortcite{Luo:2016:NAACL} observe that the students use distinct lexical items such as ``\emph{bike elements}'' and ``\emph{bicycle parts}'' to refer to the same concept. 
The student responses frequently contain expressions with little or no word overlap, such as ``\emph{the main topics of this course}'' and ``\emph{what we will learn in this class},'' when they are prompted with ``describe what you found most interesting in today's class.''
A similar phenomenon has also been observed in the news domain, where reporters use different nicknames, e.g., ``Bronx Zoo'' and ``New York Highlanders,'' to refer to the baseball team ``New York Yankees.'' 
Luo et al.,~\shortcite{Luo:2016:NAACL} report that about 80\% of the document bigrams occur only once or twice for the news domain, whereas the ratio is 97\% for student responses, suggesting the latter domain has a higher level of lexical diversity.
When source documents contain diverse expressions conveying the same meaning, it can hinder the summarization system's ability to effectively identify salient content from the source documents. 
It can also increase the summary redundancy if lexically-distinct but semantically-similar expressions are included in the summary.

Existing neural encoder-decoder models may not work well at summarizing such content with high lexical variety~\cite{rush-chopra-weston:2015:EMNLP,DBLP:journals/corr/NallapatiXZ16,Paulus:2017,see-liu-manning:2017:Long}. 
On one hand, training the neural sequence-to-sequence models requires a large amount of parallel data. 
The cost of annotating gold-standard summaries for many domains such as student responses can be prohibitive.
Without sufficient labelled data, the models can only be trained on automatically gathered corpora, where an instance often includes a news article paired with its title or a few highlights.
On the other hand, the summaries produced by existing neural encoder-decoder models are far from perfect. 
The summaries are mostly extractive with minor edits~\cite{see-liu-manning:2017:Long}\footnote{
See et al.~\shortcite{see-liu-manning:2017:Long} suggest that the pointer-generator model can copy 35\% of summary sentences from the source documents.
Similar findings are also reported by Liao et al. \shortcite{Liao:2018}, where 99.6\% of the summary unigrams, 95.2\% of bigrams, and 87.2\% of trigrams generated by the pointer-generator networks appear in the source texts.
}, contain repetitive words and phrases~\cite{Suzuki:2017} and may not accurately reproduce factual details~\cite{Cao:2018,Song:2018}.
We examine the performance of a state-of-the-art neural summarization model in Section \S\ref{sec:extrinsc}.


In this work, we propose to augment the integer linear programming (ILP)-based summarization framework with a low-rank approximation of the co-occurrence matrix, and further evaluate the approach on a broad range of datasets exhibiting high lexical diversity.
The ILP framework, being extractive in nature, has demonstrated considerable success on a number of summarization tasks~\cite{Gillick:2009:NAACL,Kirkpatrick:2011}.
It generates a summary by selecting a set of sentences from the source documents.
The sentences shall maximize the coverage of important source content, while minimizing the redundancy among themselves.
At the heart of the algorithm is a sentence-concept co-occurrence matrix, used to determine if a sentence contains important concepts and whether two sentences share the same concepts. 
We introduce a low-rank approximation to the co-occurrence matrix and optimize it using the proximal gradient method.
The resulting system thus allows different sentences to share co-occurrence statistics.
For example, ``The activity with the \textit{bicycle parts}" will be allowed to partially contain ``\textit{bike elements}" although the latter phrase does not appear in the sentence. 
The low-rank matrix approximation provides an effective way to implicitly group lexically-diverse but semantically-similar expressions.
It can handle out-of-vocabulary expressions and domain-specific terminologies well, hence being a more principled approach than heuristically calculating similarities of word embeddings.

Our research contributions of this work include the following.

\begin{itemize}
\item We present a novel ILP framework to summarize documents contributed by multiple authors and exhibiting high lexical variety. 
The framework is evaluated on eight different datasets, ranging from student course responses to product reviews and news articles.
To the best of our knowledge, our work is one of the few summarization studies assessing the proposed approach on a broad spectrum of datasets.
\vspace{0.1in}
\item Through extensive experiments, we show that the new ILP framework compares favorably to both extractive baselines and a state-of-the-art neural abstractive summarization system.
We further investigate the properties of both our system and various datasets to understand when and why the proposed approach is effective at summarizing content with high lexical variety.
\end{itemize}

In the following sections we first present a thorough review of the related work (\S\ref{sec:related_work}), then introduce our ILP summarization framework (\S\ref{sec:ilp}) with a low-rank approximation of the co-occurrence matrix optimized using the proximal gradient method (\S\ref{sec:our_approach}). 
Experiments are performed on a collection of eight datasets (\S\ref{sec:data}) containing student responses to post-class reflective questions, product reviews, peer reviews, and news articles.
Intrinsic evaluation (\S\ref{subsec:intrinsic}) shows that the low-rank approximation algorithm can effectively group distinct expressions used in similar semantic context.
For extrinsic evaluation (\S\ref{sec:extrinsc}) our proposed framework obtains competitive results in comparison to state-of-the-art summarization systems.
Finally, we conduct comprehensive studies analyzing the characteristics of the datasets and suggest critical factors that affect the summarization performance (\S\ref{chapter:sec:factors}).

\newpage

\section{Related Work}
\label{sec:related_work}

Extractive summarization has undergone great development over the past decades.
It focuses on extracting relevant sentences from a single document or a cluster of documents related to a particular topic.
Various techniques have been explored, including
maximal marginal relevance~\cite{Carbonell:1998}, 
submodularity~\cite{Lin:2010:NAACL}, 
integer linear programming~\cite{Gillick:2008,Almeida:2013,Li:2013,Li:2014:EMNLP,Durrett:2016:ACL}, 
minimizing reconstruction error~\cite{He:2012}, 
graph-based models~\cite{Erkan:2004,Radev:2004,cohan-goharian:2015:EMNLP,cohan2017scientific},
determinantal point processes~\cite{Kulesza:2012}, neural networks and reinforcement learning~\cite{Yasunaga:2017,narayan-cohen-lapata:2018:N18-1}
among others. 
Nonetheless, most studies are bound to a single dataset and few approaches have been evaluated in a cross-domain setting.
In this paper, we propose an enhanced ILP framework and evaluate it on a broad range of datasets.
We present an in-depth analysis of the dataset characteristics derived from both source documents and reference summaries to understand how domain-specific factors may affect the applicability of the proposed approach.

Neural summarization has seen promising improvements in recent years with encoder-decoder models~\cite{rush-chopra-weston:2015:EMNLP,DBLP:journals/corr/NallapatiXZ16}.
The encoder condenses the source text to a dense vector, whereas the decoder unrolls the vector to a summary sequence by predicting one word at a time.
A number of studies have been proposed to 
deal with out-of-vocabulary words~\cite{see-liu-manning:2017:Long},
improve the attention mechanism~\cite{Chen:2016,Zhou:2017,Tan:2017},
avoid generating repetitive words~\cite{see-liu-manning:2017:Long,Suzuki:2017},
adjust summary length~\cite{Kikuchi:2016}, 
encode long text~\cite{celikyilmaz-EtAl:2018:N18-1,N18-2097-Cohan-2018}
and improve the training objective~\cite{Ranzato:2016,Paulus:2017,Guo:2018:ACL}.
To date, these studies focus primarily on single-document summarization and headline generation.
This is partly because training neural encoder-decoder models requires a large amount of parallel data, yet the cost of annotating gold-standard summaries for most domains can be prohibitive. 
We validate the effectiveness of a state-of-the-art neural summarization system~\cite{see-liu-manning:2017:Long} on our collection of datasets and report results in \S\ref{sec:extrinsc}.

In this paper we focus on the integer linear programming-based summarization framework and propose enhancements to it to summarize text content with high lexical diversity.
The ILP framework is shown to perform strongly on extractive summarization~\cite{Gillick:2009:NAACL,martins-smith:2009:ILPNLP,Kirkpatrick:2011}.
It produces an optimal selection of sentences that (i) maximize the coverage of important concepts discussed in the source, (ii) minimize the redundancy in pairs of selected sentences, and (iii) ensure the summary length does not exceed a limit.
Previous work has largely focused on improving the estimation of concept weights in the ILP framework~\cite{Galanis:2012,qian-liu:2013:EMNLP2,boudin2015concept,li-liu-zhao:2015:NAACL-HLT1,Durrett:2016:ACL}.
However, distinct lexical items such as ``bike elements'' and ``bicycle parts'' are treated as different concepts and their weights are not shared.
In this paper we overcome this issue by proposing a low-rank approximation to the sentence-concept co-occurrence matrix to intrinsically group lexically-distinct but semantically-similar expressions; they are considered as a whole when maximizing concept coverage and minimizing redundancy.

Our work is also different from the traditional approaches using dimensionality reduction techniques such as non-negative matrix factorization (NNMF) and latent semantic analysis (LSA) for summarization~\cite{wang2008multi,lee2009automatic,conroy-EtAl:2013:MultiLing,conroy-davis:2015:VSM-NLP,wang2016low}. 
In particular, Wang et al.~\shortcite{wang2008multi} use NNMF to group sentences into clusters;
Conroy et al.~\shortcite{conroy-EtAl:2013:MultiLing} explore NNMF and LSA to obtain better estimates of term weights;
Wang et al.~\shortcite{wang2016low} use low-rank approximation to cast sentences and images to the same embedding space. 
Different from the above methods, our proposed framework focuses on obtaining a low-rank approximation of the co-occurrence matrix embedded in the ILP framework, so that diverse expressions can share co-occurrence frequencies. 
Note that out-of-vocabulary expressions and domain-specific terminologies are abundant in our datasets, therefore simply calculating the lexical overlap~\cite{rus:2013} or cosine similarity of word embeddings~\cite{goldberg2014word2vec} cannot serve our goal well.

This manuscript extends our previous work on summarizing student course responses~\cite{Luo:2015:EMNLP,Luo:2016:COLING,Luo:2016:NAACL} submitted after each lecture via a mobile app named CourseMIRROR~\cite{Luo:2015:demo,Fan:2015,Fan:2017:IUI}.
The students are asked to respond to reflective prompts such as ``describe what you found most interesting in today's class'' and ``describe what was confusing or needed more detail.''
For large classes with hundreds of students, it can be quite difficult for instructors to manually analyze the student responses, hence the help of automatic summarization.
Our extensions of this work are along three dimensions:
(i) we crack the ``black-box'' of the low-rank approximation algorithm to understand if it indeed allows lexically-diverse but semantically-similar items to share co-occurrence statistics;
(ii) we compare the ILP-based summarization framework with state-of-the-art baselines, including a popular neural encoder-decoder model for summarization;
(iii) we expand the student feedback datasets to include responses collected from materials science and engineering, statistics for industrial engineers, and data structures. We additionally experiment with reviews and news articles.
Analyzing the unique characteristics of each dataset allows us to identify crucial factors influencing the summarization performance.

With the fast development of Massive Open Online Courses (MOOC) platforms, more attention is being dedicated to analyzing educationally-oriented language data.
These studies seek to identify student leaders from MOOC discussion forums~\cite{Moon:2014},
perform sentiment analysis on student discussions~\cite{Wen:2014},
improve student engagement and reducing student retention~\cite{Wen:2014:ICWSM,Rose:2014}, and
using language generation techniques to automatically generate feedback to students~\cite{gkatzia-EtAl:2013:ENLG}.
Our focus of this paper is to automatically summarizing student responses so that instructors can collect feedback in a timely manner.
We expect the developed summarization techniques and result analysis will further summarization research in similar text genres exhibiting high lexical variety.

\section{ILP Formulation}
\label{sec:ilp}

Let $\mathcal{D}$ be a set of documents that consist of $M$ sentences in total. Let $y_j \in \{0,1\}$, $j = \{1,\cdots, M\}$ indicate if a sentence $j$ is selected ($y_j = 1$) or not ($y_j = 0$) in the summary. 
Similarly, let $N$ be the number of unique concepts in $\mathcal{D}$. 
$z_i \in \{0,1\}$, $i = \{1, \cdots, N\}$ indicate the appearance of concepts in the summary. 
Each concept $i$ is assigned a weight of $w_i$, often measured by the number of sentences or documents that contain the concept.
The ILP-based summarization approach~\cite{Gillick:2009:NAACL} searches for an optimal assignment to the sentence and concept variables so that the selected summary sentences maximize coverage of important concepts.
The relationship between concepts and sentences is captured by a co-occurrence matrix $A \in \mathbb{R}^{N \times M}$, where $A_{ij}=1$ indicates the $i$-th concept appears in the $j$-th sentence, and $A_{ij} = 0$ otherwise. In the literature, bigrams are frequently used as a surrogate for concepts~\cite{Gillick:2008,Kirkpatrick:2011}. 
We follow the convention and use `concept' and `bigram' interchangeably in this paper.

Two sets of linear constraints are specified to ensure the ILP validity:
(1) a concept is selected if and only if at least one sentence carrying it has been selected (Eq.~\ref{eqn:orig_ilp_bigram}), 
and (2) all concepts in a sentence will be selected if that sentence is selected (Eq.~\ref{eqn:orig_ilp_sentence}).
Finally, the selected summary sentences are allowed to contain a total of $L$ words or less (Eq.~\ref{eqn:orig_ilp_length}).

\begin{align}
\max _{\boldsymbol{y},\boldsymbol{z}} & \textstyle \quad \sum_{i=1}^{N} w_i z_i
\label{eqn:orig_ilp_obj}\\[0.6em]
s.t. & \textstyle \quad \sum_{j=1}^{M} A_{ij} \, y_j \ge z_i 
\label{eqn:orig_ilp_bigram}\\[0.3em]
& \quad A_{ij} \, y_j \le z_i 
\label{eqn:orig_ilp_sentence}\\[0.3em]
& \textstyle \quad \sum_{j=1}^{M} l_j y_j \le L 
\label{eqn:orig_ilp_length}\\[0.3em]
& \quad y_j \in \{0,1\}
\label{eqn:orig_y}\\[0.3em]
& \quad z_i \in \{0,1\}
\label{eqn:orig_z}
\end{align}

\noindent The above ILP can be transformed to matrix representation:

\begin{align}
\max _{\boldsymbol{y},\boldsymbol{z}} & \quad \boldsymbol{w}^\top\boldsymbol{z}
\label{eqn:ilp_obj}\\[0.6em]
s.t. & \quad \boldsymbol{A} \, \boldsymbol{y} \ge \boldsymbol{z} 
\label{eqn:ilp_bigram}\\[0.3em]
& \quad \boldsymbol{A} \, \text{diag}(\boldsymbol{y}) \le \boldsymbol{Z}  \label{eqn:ilp_sentence}\\[0.3em]
& \quad \boldsymbol{\eta}^{\top}\boldsymbol{y} \le L 
\label{eqn:ilp_length}\\[0.3em]
& \quad \boldsymbol{y} \in \{0,1\}^{M} 
\label{eqn:ilp_sent_var}\\[0.3em]
& \quad \boldsymbol{z} \in \{0,1\}^{N} 
\label{eqn:ilp_concept_var}
\end{align}

We use boldface letters to represent vectors and matrices.
$\boldsymbol{Z} = [\boldsymbol{z}, ..., \boldsymbol{z}] \in \mathbb{R}^{N \times M}$ is an auxiliary matrix created by horizontally stacking the concept vector $\boldsymbol{z} \in \mathbb{R}^N$ $M$ times.
Constraint set~(Eq.~\ref{eqn:ilp_sentence}) specifies that a sentence is selected indicates that all concepts it carries have been selected.
It corresponds to $N \times M$ constraints of the form $\textstyle A_{i,j} \, y_j \le z_i$, where $i \in [N], j \in [M]$.

As far as we know, this is the first-of-its-kind matrix representation of the ILP framework. It clearly shows the two important components of this framework, including 1) the concept-sentence co-occurrence matrix $\boldsymbol{A}$, and 2) concept weight vector $\boldsymbol{w}$. Existing work focus mainly on generating better estimates of concept weights ($\boldsymbol{w}$), while we focus on improving the co-occurrence matrix $\boldsymbol{A}$.

\section{Our Approach}
\label{sec:our_approach}

Because of the lexical diversity problem, we suspect the co-occurrence matrix $\boldsymbol{A}$ may not establish a faithful correspondence between sentences and concepts.
A concept may be conveyed using multiple bigram expressions; however, the current co-occurrence matrix only captures a binary relationship between sentences and bigrams.
For example, we ought to give partial credit to ``bicycle parts''  given that a similar expression ``bike elements'' appears in the sentence.
Domain-specific synonyms may be captured as well. 
For example, the sentence ``I tried to follow along but I couldn't \textit{grasp the} concepts'' is expected to partially contain the concept ``understand the'', although the latter did not appear in the sentence.

The existing matrix $\boldsymbol{A}$ is highly sparse. Only 3.7\% of the entries are non-zero in the student response data sets on average (\S\ref{sec:data}).
We therefore propose to \emph{impute} the co-occurrence matrix by filling in missing values (i.e., matrix completion).
This is accomplished by approximating the original co-occurrence matrix using a low-rank matrix. The low-rankness encourages similar concepts to be shared across sentences. 

The ILP with a low-rank approximation of the co-occurrence matrix can be formalized as follows.

\begin{align}
\max _{\boldsymbol{y},\boldsymbol{z}} & \quad \boldsymbol{w}^\top\boldsymbol{z}
\label{eqn:ilp_obj_new}\\[0.6em]
s.t. & \quad \boldsymbol{\hat{A}} \, \boldsymbol{y} \ge \boldsymbol{z} 
\label{eqn:ilp_bigram_new}\\[0.3em]
& \quad \boldsymbol{\hat{A}} \, \text{diag}(\boldsymbol{y}) \le \boldsymbol{Z}  \label{eqn:ilp_sentence_new}\\[0.3em]
& \quad \boldsymbol{\eta}^{\top}\boldsymbol{y} \le L 
\label{eqn:ilp_length_new}\\[0.3em]
& \quad \boldsymbol{y} \in \{0,1\}^{M} 
\label{eqn:ilp_sent_var_new}\\[0.3em]
& \quad \boldsymbol{z} \in [0,1]^{N} 
\label{eqn:ilp_concept_var_new}
\end{align}

The low-rank approximation process makes two notable changes to the existing ILP framework. 
\begin{itemize}
    \item It extends the domain of $A_{ij}$ from binary to a continuous scale $[0,1]$ (Eq.~\ref{eqn:ilp_bigram_new} and Eq.~\ref{eqn:ilp_sentence_new}), which offers a better sentence-level semantic representation. 
    \item The binary concept variables ($z_i$) are also relaxed to continuous domain $[0,1]$ (Eq.~\ref{eqn:ilp_concept_var_new}), which allows the concepts to be ``partially'' included in the summary.
\end{itemize}


Concretely, given the co-occurrence matrix $\boldsymbol{A} \in \mathbb{R}^{N \times M}$, we aim to find a low-rank matrix $\boldsymbol{\hat{A}} \in \mathbb{R}^{N \times M}$ whose values are close to $\boldsymbol{A}$ at the observed positions. Our objective function is
\begin{align}
\label{eqn:mc_obj}
\displaystyle \min_{\hat{A} \in \mathbb{R}^{N \times M}} \frac{1}{2}\sum_{(i,j) \in \Omega} (A_{ij} - \hat{A}_{ij})^2 + \lambda \left\| \hat{A} \right\|_{*} ,
\end{align}
 
\noindent where $\Omega$ represents the set of observed value positions. $\|\hat{A}\|_*$ denotes the trace norm of $\hat{A}$, i.e., $\|\hat{A}\|_* = \sum_{i=1}^r \sigma_i$, where $r$ is the rank of $\hat{A}$ and $\sigma_i$ are the singular values. 
By defining the following projection operator $P_{\Omega}$, 
\begin{align}
\label{eqn:mc_proj_op}
\displaystyle 
[P_{\Omega}(\hat{A})]_{ij} = \left\{
     \begin{array}{lr}
       \hat{A}_{ij} & (i,j) \in \Omega\\
       0 & (i,j) \notin \Omega
     \end{array}
   \right.
\end{align}

\noindent our objective function~(Eq.~\ref{eqn:mc_obj}) can be succinctly represented as 
\begin{align}
\label{eqn:new_obj}
\min_{\hat{A} \in \mathbb{R}^{N \times M}} \frac{1}{2} \| P_{\Omega}(A) - P_{\Omega}(\hat{A}) \|_{F}^2 + \lambda \| \hat{A} \|_{*} ,
\end{align}

\noindent where $\| \cdot \|_F$ denotes the Frobenius norm.

Following Mazumder et al.~\shortcite{Mazumder:2010}, we optimize Eq. \ref{eqn:new_obj} using the proximal gradient descent algorithm. The update rule is 
\begin{small}
\begin{equation}
    \hat{A}^{(k+1)} = \mbox{prox}_{\lambda \rho_k} \Big( \hat{A}^{(k)} + \rho_k \big( P_{\Omega}(A) - P_{\Omega}(\hat{A}) \big) \Big) ,
\end{equation}
\end{small}

\noindent where $\rho_k$ is the step size at iteration \emph{k} and the proximal function $\mbox{prox}_{t}(\hat{A})$ is defined as the singular value soft-thresholding operator, $\mbox{prox}_{t}(\hat{A}) = U \cdot \mbox{diag}((\sigma_i - t)_+) \cdot V^\top$, where $\hat{A} = U \mbox{diag}(\sigma_1,\cdots,\sigma_r)V^\top$ is the singular value decomposition (SVD) of $\hat{A}$ and $(x)_+ = \max(x, 0)$.

Since the gradient of $\frac{1}{2} \| P_{\Omega}(A) - P_{\Omega}(\hat{A}) \|_{F}^2$ is Lipschitz continuous with $L = 1$ ($L$ is the Lipschitz continuous constant), we follow Mazumder et al.~\shortcite{Mazumder:2010} to choose fixed step size $\rho_k = 1$, which has a provable convergence rate of $O(1/k)$, where $k$ is the number of iterations.

\section{Datasets}
\label{sec:data}

To demonstrate the generality of the proposed approach, we consider three distinct types of corpora, ranging from student response data sets from four different courses to three sets of reviews to one benchmark of news articles.  The corpora are summarized in Table~\ref{table:existing_dataset}. 

\begin{table*}[htb]
\begin{minipage}{\textwidth}
\begin{center}
\begin{small}
\begin{tabular}{lccccc}
\hline
\hline
& \multicolumn{5}{c}{Statistics} \\\cline{2-6}
& Tasks & Docs/task & WC/task  & WC/sen& Length \\
\hline
Student response (Eng)* & 36 & 49 & 375.4 & 9.1 & 30\\
Student response (Stat2015)* & 44 & 39 & 233.1 & 6.0 & 15  \\
Student response (Stat2016)* & 48 & 42 & 149.3 & 4.3 & 13 \\
Student response (CS2016) & 46 & 22 & 223.1 & 8.8 & 16 \\\hline
Reviews (camera)  & 3 & 18 & 1927.0 & 22.7 & 216\\
Reviews (movie)  & 3 & 18 & 8014.0 & 24.4 & 242\\
Reviews (peer)  & 3 & 18 & 1543.7 & 19.2 & 190\\\hline
News articles (DUC04)* & 50 & 10 & 5171.6 & 22.4 & 105 \\\hline
\hline
\end{tabular}
\end{small}

\end{center}
\end{minipage}
\caption{Selected summarization data sets. Publicly available data sets are marked with an asterisk (*). The statistics involve the number of summarization tasks (Tasks), average number of documents per task (Docs/task), average word count per task (WC/task), average word count per sentence (WC/sen), and average number of words in human summaries (Length).}
\label{table:existing_dataset}
\end{table*}

\vspace{0.2in}
\noindent\textbf{Student responses.}
Research has explored using reflection prompts/muddy cards/one-minute papers to promote and collect reflections from students~\cite{Wilson:1986,Mosteller:1989,Harwood:1996}. However, it is expensive and time consuming for humans to summarize such feedback. It is therefore desirable to automatically summarize the student feedback produced in online and offline environments, although it is only recently that a data collection effort to support such research has been initiated~\cite{Fan:2015,Luo:2015:demo}. In our data, one particular type of student response is considered, named  ``\textit{reflective feedback}'' \cite{boud:2013}, which has been shown to enhance interaction between instructors and students by educational researchers \cite{vandenBoom:2004,Menekse:2011}. More specifically, students are presented with the following prompts after each lecture and  asked to provide responses: 1) ``{describe what you found most interesting in today's class},'' 
2) ``{describe what was confusing or needed more detail},'' 
and 3) ``{describe what you learned about how you learn}.''
These open-ended prompts are carefully designed to encourage students to self-reflect, allowing them to ``recapture experience, think about it and evaluate it"~\cite{boud:2013}.

\begin{table}[t]
\begin{small}
\begin{minipage}{\textwidth}
\centering
\begin{tabular}{ l}
\hline

\rule{0pt}{3ex}\textbf{Prompt}\\
Describe what you found most interesting in today's class\\
\hline
\rule{0pt}{3ex}\textbf{Student Responses}\\
S1: \hilightr{Guilt analogy}$^r$ \\ 
S2: \hilightg{Error bounding}$^g$ is interesting and useful\\
S3: the idea of c and \hilightg{finding that error}$^g$ looked great to me\\
S4: You stated that the concept of the \hilightg{error boundary}$^g$ is abstract however \\
\quad \quad i got it very well \\
S5: determining the probability of the \hilightg{error}$^g$ while rejecting \hilightb{ho}$^b$.\\
S6: The process of \hilighty{hypothesis testing}$^y$\\
S7: \hilightg{Determining the critical value for error}$^g$\\
S8: \hilightb{H1 and Ho conditions}$^b$\\
S9: Baydogan finally check the students in the class. But i think it must be in every \\
\quad \quad lecture even in the PS\\
S10: Testing whether the information we have is true or not with \hilighty{hypothesis testing}$^y$ \\
\quad \quad method was interesting\\
...\\
\hline
\rule{0pt}{3ex}\textbf{Human Summary 1}\\
- \hilighty{Hypothesis testing (in general)}$^y$ \\ 
- \hilightg{Error bounding}$^g$\\
- \hilightr{Guilt analogy helpful}$^r$\\
- \hilightb{Conditions for H1 and H0}$^b$\\
- \hilightm{Good use of examples}$^m$\\
\hline
\end{tabular}
\end{minipage}
\caption{
Example prompt, student responses, and one human summary. `S1'--`S10' are student IDs.
The summary and phrase highlights are manually created by annotators.
Phrases that bear the same color belong to the same issue.
The superscripts of the phrase highlights are imposed by the authors to differentiate colors when printed in grayscale (y:\hilighty{yellow}, g:\hilightg{green}, r:\hilightr{red}, b:\hilightb{blue}, and m:\hilightm{magenta}).
}
\label{table:example}
\end{small}
\end{table}

To test generality, we gathered student responses from four different courses, as shown in Table~\ref{table:existing_dataset}. The first one was collected by Menekse et al.~\shortcite{Menekse:2011} using paper-based surveys from an introductory materials science and \underline {eng}ineering class (henceforth {\bf Eng}) taught in a major U.S. university, and a subset is made public by us~\cite{Luo:2015:EMNLP}, available at the link: \url{http://www.coursemirror.com/download/dataset}. The remaining three courses are collected by us using a mobile application, CourseMIRROR~\cite{Luo:2015:demo,Fan:2015} and then the reference summaries for each course are created by human annotators with the proper background. The human annotators are allowed to create abstract summaries using their own words in addition to selecting phrases directly from the responses. While the $2^{nd}$ and $3^{rd}$ data sets are from the same course, \underline{Stat}istics for Industrial Engineers, they were taught in \underline{2015} and \underline{2016} respectively (henceforth {\bf Stat2015} and {\bf Stat2016}), at the Bo\v{g}azi\c{c}i University in Turkey.\footnote{Publicly available at \url{http://www.coursemirror.com/download/dataset2}~\cite{Luo:2016:COLING}} The course was taught in English while the official language is Turkish. The last one is from a fundamental undergraduate \underline{C}omputer \underline{S}cience course (data structures) at a local U.S. university taught in \underline{2016} (henceforth {\bf CS2016}).

Another reason we choose the student responses is that we have advanced annotation allowing us to perform an intrinsic evaluation to test whether the low-rank approximation does capture similar concepts or not. An example of the annotation is shown in Table~\ref{table:example}, where phrases in the student responses that are semantically the same as the summary phrases are highlighted with the same color by human annotators. For example, ``error bounding" (S2), ``error boundary" (S4), ``finding that error" (S3), and ``determining the critical value for error" (S7) are semantically equivalent to ``Error bounding" in the human summary. Details of the intrinsic evaluation are introduced in \ref{subsec:intrinsic}.

\vspace{0.2in}
\noindent\textbf{Product and peer reviews.}
The review data sets are provided by Xiong and Litman~\shortcite{xiong-litman:2014:Coling}, consisting of 3 categories. 
The first one is a subset of product reviews from a widely used data set in review opinion mining and sentiment analysis, contributed by Jindal and Liu~\shortcite{jindal2008opinion}. In particular, it randomly sampled 3 set of reviews from a representative product (digital camera), each with 18 reviews from an individual product type (e.g. ``summarizing 18 camera reviews for Nikon D3200"). 
The second one is movie reviews crawled from IMDB.com by the authors themselves. 
The third one is peer reviews collected in a college-level history class from an
online peer-review reciprocal system, SWoRD \cite{cho2008machine}. 
The average number of sentences per review set is 85 for camera reviews, 328 for movie reviews and 80 for peer review; the average number of words per sentence in the camera, movie, and peer reviews are 23, 24 and 19, respectively. 
The human summaries were collected in the form of online surveys (one survey per domain) hosted by Qualtrics. Each human summary contains 10 sentences from users' reviews. Example movie reviews are shown in Table~\ref{tab:example_movie_review}.

\begin{table}[t]
\begin{small}
\begin{minipage}{\textwidth}
\centering
\begin{tabular}{ l}
\hline
\rule{0pt}{3ex}``Forrest Gump" is one of the best movies of all time, guaranteed.\\
I just love this movie.\\
It truly is amazing...\\
What an amazing story and moving meaning.\\
I really just love this movie and it has such a special place in my heart.\\
And anyone who hasn't seen it or who thinks that don't like it I seriously suggest seeing\\
\quad it or seeing it again.\\
That movie teaches you so much about life and the meaning of it.\\
This is one masterpiece of a movie that will not be forgotten about in a long time.\\
This is a powerful yet charming movie; fun for its special effects and profound in how it \\
\quad keeps you thinking long after it's over.\\
It may change your lifeOne hell of a movie; it will be close to my heart forever!\\
It is something to mull over for a long time.\\
The performances are just so unforgettable and never get out of your head.\\
I've watched the movie about once every two years since then.\\
The lines are so memorable, touching, and sometimes hilarious.We have Forrest Gump \\
\quad (Tom Hanks), not the sharpest tool in the box, his I.Q.\\
Well done, well acted, and well directed to pythagorean procision. A++\\
This story is beautiful and will inspire everyone to go the distance and see the world \\
\quad like Forrest did and will never give up on their dreams.10/10\\
A++\\
You 'd be a fool to miss it.Bottom Line : 4 out of 4 (own this movie)\\
\hline
\end{tabular}
\end{minipage}
\caption{Example movie reviews.}
\label{tab:example_movie_review}
\end{small}
\end{table}

\noindent\textbf{News articles.}
Most summarization work focuses on news documents, as driven by the Document Understanding Conferences (DUC) and Text Analysis Conferences (TAC). For comparison, we select \underline{DUC} 20\underline{04}\footnote{\url{http://duc.nist.gov/duc2004/}} to evaluate our approach (henceforth {\bf DUC04}), which is widely used in the literature~\cite{Lin:2004,HONG14.1093.L14-1070,ren-EtAl:2016:COLING,takase-EtAl:2016:EMNLP2016,wang-EtAl:2016:COLING1}. It consists of 50 clusters of Text REtrieval Conference (TREC) documents, from the following collections: AP newswire, 1998-2000; New York Times newswire, 1998-2000; Xinhua News Agency (English version), 1996-2000. Each cluster contained on average 10 documents. The task is to create a short summary ($\le$ 665 bytes) of each cluster. Example news sentences are shown in Table~\ref{tab:example_news}.

\begin{table}[t]
\begin{small}
\begin{minipage}{\textwidth}
\centering
\begin{tabular}{ l}
\hline
Samaranch expressed surprise at allegations made by the IOC executive board member \\
\quad Marc Hodler of Switzerland that agents were offering to sell I.O.C. members' votes \\
\quad for payments from bidding cities.\\
Moving quickly to tackle an escalating corruption scandal, IOC leaders questioned Salt \\
\quad Lake City officials Friday in the first ever investigation into alleged vote-buying by an \\
\quad Olympic city.\\
Acting with unusual speed, the International Olympic Committee set up a special \\
\quad investigative panel that immediately summoned the organizers of the 2002 Salt Lake \\
\quad Games to address the bribery allegations. \\
It's the most serious case of alleged ethical misconduct investigated by the IOC since \\
\quad former U.S. member Robert Helmick was accused of conflict of interest in 1991.\\
This is the first time the IOC has ever investigated possible bribery by  bidding cities, \\
\quad despite previous rumors and allegations of corruption in other Olympic votes.\\
Hodler said a group of four agents, including one IOC member, have been  involved in \\
\quad promising votes for payment.\\
Samaranch Sunday ruled out taking the Games from Salt Lake City.\\
I can't be stronger in saying I don't consider it a possibility whatsoever of the games \\
\quad being withdrawn from Salt Lake City.\\
The chief investigator refused to rule out the possibility of taking the games away from \\
\quad Salt Lake City - though that scenario is considered highly unlikely.\\
\hline
\end{tabular}
\end{minipage}
\caption{Example sentences from news.}
\label{tab:example_news}
\end{small}
\end{table}

\section{Experiments}
\label{sec:experiments}

In this section, we evaluate the proposed method intrinsically in terms of whether the co-occurrence matrix after the low-rank approximation is able to capture similar concepts on student response data sets, and also extrinsically in terms of the end task of summarization on all corpora. In the following experiments, summary length is set to be the average number of words in human summaries or less. For the matrix completion algorithm, we perform grid search (on a scale of [0, 5] with stepsize 0.5) to tune the hyper-parameter $\lambda$ (Eq.~\ref{eqn:mc_obj}) with a leave-one-lecture-out (for student responses) or leave-one-task-out (for others) cross-validation.

\subsection{Intrinsic evaluation}
\label{subsec:intrinsic}

When examining the imputed sentence-concept co-occurrence matrix, we notice some interesting examples that indicate the effectiveness of the proposed approach, shown in Table~\ref{tab:example_mc}. 

\begin{table}[!ht]
\setlength{\tabcolsep}{5pt}
\renewcommand{\arraystretch}{1.1}
\begin{minipage}{\textwidth}
\begin{center}
\begin{tabular}{l | r }
\hline
\textbf{Sentence} & \textbf{Assoc. Bigrams}\\ 
\hline
\hline
\rule{0pt}{2ex}\textit{the printing} needs to better so it can  be easier to read & \textit{the graph}\\\hline
\rule{0pt}{2ex}graphs make it \textit{easier to} understand concepts & \textit{hard to}\\\hline
\rule{0pt}{2ex}the naming system for the 2 \textit{phase regions} & \textit{phase diagram}\\\hline
\rule{0pt}{2ex}I tried to follow along but I couldn't \textit{grasp the} concepts & \textit{understand the}\\\hline
\rule{0pt}{2ex}no problems except for the specific equations used to & \multirow{2}{*}{\textit{strain curves}}\\
\rule{0pt}{2ex} \quad determine properties from the stress - \textit{strain graph} & \\\hline
\rule{0pt}{2ex}why delete the first entry in the \textit{linked bag} instead of & \textit{linked list}\\
\rule{0pt}{2ex} \quad just moving the pointers from the node & \\
\rule{0pt}{2ex} \quad before the deleted node to the node after & \\\hline
\rule{0pt}{2ex}You make \textit{a movie} that romanticizes the `50's, & \textit{the film}\\
\rule{0pt}{2ex} \quad `60's and `70's, and with enough publicity and & \\
\rule{0pt}{2ex} \quad a good enough soundtrack ... & \\\hline
\rule{0pt}{2ex}\textit{U.S.} officials have said the construction ... & \textit{united states}\\\hline
\rule{0pt}{2ex}\textit{American} officials have said spy satellites ... & \textit{united states}\\\hline
\rule{0pt}{2ex}It also sought to cast \textit{Gates} as an obsessed man who & \textit{that microsoft}\\
\rule{0pt}{2ex} \quad feared the tiny Netscape Communications Corp. & \\
\rule{0pt}{2ex} \quad and its potential threat to his domination of the & \\ \rule{0pt}{2ex} \quad market for \textit{Internet browsers}, the software used to & \\ \rule{0pt}{2ex} \quad navigate the World Wide Web.  \\
\hline
\end{tabular}
\end{center}
\end{minipage}
\caption{Associated bigrams that do not appear in the sentence, but after Matrix Completion, yield a decent correlation (cell value greater than 0.9) with the corresponding sentence.
}
\label{tab:example_mc}
\end{table}

We want to investigate whether the matrix completion (MC) helps to capture similar concepts (i.e., bigrams). Recall that, if a bigram $i$ is similar to another bigram in a sentence $j$, the sentence $j$ should assign a partial score to the bigram $i$ after the low-rank approximation. For instance, ``The activity with the bicycle parts" should give a partial score to ``bike elements" since it is similar to ``bicycle parts". 
Note that, the co-occurrence matrix $\boldsymbol{A}$ measures whether a sentence includes a bigram or not. Without matrix completion, if a bigram $i$ does not appear in a sentence $j$, $A_{ij} = 0$. After matrix completion, $\hat{A}_{ij}$ ($\hat{A}$ is the low-rank approximation matrix of $A$) becomes a continuous number ranging from 0 to 1 (negative values are truncated). Therefore, $\hat{A}_{ij}>0$ does not necessarily mean the sentence contains a similar bigram, since it might also give positive scores to non-similar bigrams.
To solve this issue, we propose two different ways to test whether the matrix completion really helps to capture similar concepts.

\begin{itemize}

    \item H1.a: A bigram receives a higher partial score in a sentence that contains similar bigram(s) to it than a sentence that does not. That is, if a bigram $i$ is similar to one of bigrams in a sentence $j^+$, but not similar to any bigram in another sentence $j^-$, then after matrix completion, $\hat{A}_{ij^+} > \hat{A}_{ij^-}$. 
    
    \item H1.b: A sentence gives higher partial scores to bigrams that are similar to its own bigrams than bigrams that are different from its own. That is, if a sentence $j$ has a bigram that is similar to $i^+$, but none of its bigrams is similar to $i^-$, then, after matrix completion, $\hat{A}_{i^+j} > \hat{A}_{i^-j}$.

\end{itemize}

In order to test these two hypotheses, we need to construct gold-standard pairs of similar bigrams and pairs of different bigrams, which can be automatically obtained with the phrase-highlighting data (Table~\ref{table:example}). We first extract a candidate bigram from a phrase if and only if a single bigram can be extracted from the phrase. In this way, we discard long phrases if there are multiple candidate bigrams among them in order to avoid ambiguity as we cannot validate which of them match another target bigram. A bigram is defined as two words and at least one of them is not a stop word. We then extract every pair of candidate bigrams that are highlighted in the same color as similar bigrams. Similarly, we extract every pair of candidate bigrams that are highlighted as different colors as different bigrams. For example, ``bias reduction" is a candidate phrase, which is similar to ``bias correction" since they are in the same color. 

To test H1.a, given a bigram $i$, a bigram $i^+$ that is similar to it, and a bigram $i^-$ that is different from it, we can select the bigram $i$, and the sentence $j^+$ that contains $i^+$, and the sentence $j^-$ that contains $i^-$. We ignore $j^-$ if it contains any other bigram that is similar to $i$ to eliminate the compounded case that both similar and different bigrams are within one sentence. Note, if there are multiple sentences containing $i^+$, we consider each of them. In this way, we construct a triple $\left \langle i,j^+,j^-\right \rangle$, and test whether $\hat{A}_{ij^+} > \hat{A}_{ij^-}$. 
To test H1.b, for each pair of similar bigrams $\left \langle i,i^+\right \rangle$, and different bigrams $\left \langle i,i^-\right \rangle$, we select the sentence $j$ that contains $i$ so that we construct a triple $\left \langle i^+,i^-,j\right \rangle$, and test whether $\hat{A}_{i^+j} > \hat{A}_{i^-j}$. We also filtered out $j$ that contains similar bigram(s) to $i^-$ to remove the compounded effect.
In this way, we collected a gold-standard data set to test the two hypotheses above as shown in Table~\ref{table:intrinsic_size}. 

\begin{table}
\begin{minipage}{\textwidth}
\begin{center}
\begin{tabular}{c|c|c|c|c|c}
\hline\hline
\textbf{Corpus}	& \textbf{bigrams} & \textbf{similar pairs} & \textbf{different pairs} & \textbf{$\left \langle i,j^+,j^-\right \rangle$} & \textbf{$\left \langle i^+,i^-,j\right \rangle$} \\
	\hline
Stat2015 & 516 & 198 & 698 & 404 & 279 \\
Stat2016 & 1,673 & 638 & 1,928 & 1,188 & 228 \\
CS2016 & 613 & 168 & 412 & 235 & 46 \\
\hline\hline
\end{tabular}

\end{center}
\end{minipage}
\caption{A gold-standard data set was extracted from three student response corpora that have phrase-highlighting annotation. Statistics include: the number of bigrams, the number of pairs of similar bigrams and pairs of different bigrams, the number of tuples $\left \langle i,j^+,j^-\right \rangle$, and the number of $\left \langle i^+,i^-,j\right \rangle$. $i$ is a bigram, $j^+$ is a sentence with a bigram similar to $i$, and $j^-$ is a sentence with a bigram different from $i$. $j$ is a sentence, $i^+$ is a bigram that is similar to a bigram in $j$, and $i^-$ is a bigram that is different from any bigram in $j$.}
\label{table:intrinsic_size}
\end{table}

\begin{table}
\centering
\begin{minipage}{0.8\textwidth}
\begin{tabular}{c|c|c|c|c|c|c}
\hline\hline
   &  \multicolumn{2}{c|}{Stat2015}  & \multicolumn{2}{c|}{Stat2016}  &  \multicolumn{2}{c}{CS2016} \\\hline
\multirow{2}{*}{H1.a}   & $\hat{A}_{ij^+}$  & $\hat{A}_{ij^-}$ & $\hat{A}_{ij^+}$  & $\hat{A}_{ij^-}$  & $\hat{A}_{ij^+}$  & $\hat{A}_{ij^-}$  \\\cline{2-7}
 & 0.122$^*$ & 0.056 & 0.108$^*$ & 0.038 & 0.238$^*$ & 0.089 \\\hline
\multirow{2}{*}{H1.b}    &  $\hat{A}_{i^+j}$ & $\hat{A}_{i^-j}$ & $\hat{A}_{i^+j}$ & $\hat{A}_{i^-j}$ & $\hat{A}_{i^+j}$ & $\hat{A}_{i^-j}$ \\\cline{2-7}
 & 0.147 & 0.151 & 0.132$^*$ & 0.074  & 0.186 & 0.149\\
 \hline\hline
\end{tabular}

\end{minipage}
\caption{Hypothesise testing: whether the matrix completion (MC) helps to capture similar concepts. $^*$ means $p < 0.05$ using a two-tailed paired t-test.}
\label{table:intrinsic_res}
\end{table}

The results are shown in Table \ref{table:intrinsic_res}.  $\hat{A}_{ij^+} > \hat{A}_{ij^-}$ significantly on all three courses. That is, a bigram does receive a higher partial score in a sentence that contains similar bigram(s) to it than a sentence that does not. Therefore, H1.a holds. For H1.b, we only observe $\hat{A}_{i^+j} > \hat{A}_{i^-j}$ significantly on Stat2016 and there is no significant difference between $\hat{A}_{i^+j}$ and $\hat{A}_{i^-j}$ on the other two courses. First, the gold-standard data set is still small in the sense that only a limited portion of bigrams in the entire data set are evaluated. Second, the assumption that phrases annotated by different colors are not necessarily unrelated is too strong. For example, ``hypothesis testing" and ``H1 and Ho conditions" are in different colors in the example of Table~\ref{table:example}, but one is a subtopic of the other. An alternative way to evaluate the hypothesis is to let humans judge whether two bigrams are similar or not, which we leave for future work. Third, the gold standards are pairs of semantically similar bigrams, while matrix completion captures bigrams that occurs in a similar context, which is not necessarily equivalent to semantic similarity. For example, the sentence ``graphs make it \textit{easier to} understand concepts" in Table~\ref{table:intrinsic_res} is associated with ``hard to".

\begin{table}
\begin{center}
\begin{footnotesize}
\begin{tabular}{l l | l | l| l | l|c}
\hline\hline
& \multicolumn{1}{l|}{\textbf{System}} & \textbf{R-1} & \textbf{R-2} & \textbf{R-SU4} & \textbf{R-L} & \textbf{Human Preference}\\
\hline
Eng         &  MEAD     &.192$^*$               &.052$^*$               &.053$^*$               &.183$^*$           & -\\
            &  LexRank  &.303$^*$               &.093                   &.098$^*$               &.278$^*$           & -\\
            &  SumBasic &.387                   &.090$^*$               &\bf.153                &.370               & 26.9\%\\
            &  PGN      &.237$^*$               &.047$^*$               &.066$^*$               &.218$^*$           & - \\
            &  ILP      &.364$^*$               &.123                   &.135                   &.347$^*$           & 24.1\% \\
            &  ILP+MC   &\bf\underline{.392}    &\bf\underline{.130}    &\underline{.150}       &\bf\underline{.380}&  \bf\underline{29.4\%} \\\hline
Stat2015    &  MEAD     &.225$^*$               &.073$^*$               &.065$^*$               &.218$^*$           & -\\
            &  LexRank  &.334$^*$               &.154                   &.134$^*$               &.326$^*$           & -\\
            &  SumBasic &\bf.457$^*$            &\bf.193                &\bf.223$^*$            &.444$^*$           & \bf 30.7\%\\
             &  PGN  &  .253$^*$    &  .096$^*$   &  .087$^*$   &  .241$^*$ & - \\
            &  ILP      &.405                   &.186                   &.165                   &\bf.397            & \ 29.2\%$^*$ \\
            &  ILP+MC   &.401                   &.183                   &\underline{.173}       &.391               &  \underline{29.6\%} \\\hline
Stat2016    &  MEAD     &.364$^*$               &.172$^*$               &.140$^*$               &.347$^*$           & -\\
            &  LexRank  &.397$^*$               &.191                   &.168                   &.384$^*$           &-\\
            &  SumBasic &\bf.554$^*$            &\bf.295$^*$            &\bf.292$^*$            &\bf.538$^*$        & \bf 32.9\%\\
             &  PGN  &  .277$^*$   &  .106$^*$   &  .092$^*$   &  .269$^*$& -  \\
            &  ILP      &.482                   &.262$^*$               &.215                   &.468               & 29.1\% \\
            &  ILP+MC   &.457                   &.214                   &.198                   &.441               &  28.0\% \\\hline
CS2016      &  MEAD     &.221$^*$               &.056$^*$               &.064$^*$               &.207$^*$           & -\\
            &  LexRank  &.285$^*$               &.085$^*$               &.091$^*$               &.272$^*$           & -\\
            &  SumBasic &\bf.408                &.141                   &.164                   &\bf.393            & 31.5\%\\
            & PGN  &  .204$^*$  &  .057$^*$  &  .056$^*$  &  .192$^*$& -  \\
            &  ILP      &.374                   &.141                   &.137$^*$               &.356$^*$           & \ 24.4\%$^*$ \\
            &  ILP+MC   &\underline{.398}       &\bf\underline{.154}    &\bf\underline{.158}    &\underline{.383}   &\bf\underline{32.7\%} \\\hline
camera      &  MEAD     &\bf.475                &\bf.207                &\bf.211                &.421               & -\\
            &  LexRank  &.439                   &.181                   &.183                   &.398               & -\\
            &  SumBasic &\bf.475                &.168                   &.196                   &\bf.439            & \ 23.9\%$^*$\\
             &  PGN  &  .106$^*$ &  .050  &  .011$^*$  &  .101$^*$& -  \\
            &  ILP      &.457                   &.165                   &.181                   &.427               & \bf36.9\% \\
            &  ILP+MC   &.447                   &.157                   &.176                   &.418               & 32.5\% \\\hline
movie       &  MEAD     &.394                   &.131                   &.146                   &.341               & -\\
            &  LexRank  &.434$^*$               &\bf.147                &\bf.186                &.386               & -\\
            &  SumBasic &\bf.441                &.098                   &.176                   &.401               & \ 27.6\%$^*$\\
             &  PGN  &  .108$^*$  &  .047$^*$ &  .011$^*$  &  .101$^*$ & - \\
            &  ILP      &.435                   &.091$^*$               &.167                   &.397               & \bf36.9\% \\
            &  ILP+MC   &\underline{.436}       &\underline{.106}       &\underline{.169}       &\bf\underline{.409}& 21.8\% \\\hline
peer        &  MEAD     &.469                   &.242                   &\bf.212                &.436               & -\\
            &  LexRank  &.444                   &.196                   &.170                   &.417               & -\\
            &  SumBasic &.473                   & .154$^*$              &.199                   &.441               & 23.3\%\\
             &  PGN  &  .161   &  .098  &  .029   &  .155 & - \\
            &  ILP      &.466                   &.199                   &.183                   &.445               &\bf34.4\% \\
            &  ILP+MC   &\bf\underline{.491}    &\bf\underline{.261}    &\underline{.195}       &\bf\underline{.469}& 22.2\% \\\hline
DUC04       &  MEAD     &.352                   &.076                   &.117                   &.299               & -\\
            &  LexRank  &.354                   &.076                   &.118                   &.308               & -\\
            &  SumBasic &.364$^*$               &.066                   &.117                   &.326$^*$           & \ 24.9\%$^*$\\
            &  PGN  &  .155$^*$  &  .025$^*$   &  .023$^*$   &  .137$^*$& - \\
            &  ILP      &\bf.377$^*$            &\bf.092$^*$            &\bf.126$^*$            &\bf.333$^*$        & \ 27.3\%$^*$ \\
            &  ILP+MC   &.342                   &.072                   &.109                   &.308               &\bf\underline{31.1\%}\\
\hline
\end{tabular}
\end{footnotesize}

\end{center}
\vspace{-10pt}
\caption{Summarization results evaluated by ROUGE and human judges. 
Best results are shown in {\bf bold} for each data set. $^*$ indicates that the performance difference with ILP+MC is statistically significant (p $<$ 0.05) using a two-tailed paired t-test. \underline{Underline} means that ILP+MC is  better than ILP.
}
\label{tab:results_ilp}
\end{table}

\begin{table}
\begin{center}
\begin{minipage}{0.8\textwidth}
\begin{tabular}{r | l | l | l| l }
\hline\hline
& \textbf{R-1} & \textbf{R-2} & \textbf{R-SU4} & \textbf{R-L}\\
\hline
Human Annotator 1 & .418 & .105 & - & .431 \\
Human Annotator 2 & .412 & .086 & - & .409 \\
Human Annotator 3 & .410 & .090 & - & .434 \\
Human Annotator 4 & .406 & .098 & - & .420 \\
Human Annotator 5 & .404 & .107 & - & .425 \\
Human Annotator 6 & .393 & .097 & - & .406 \\
Human Annotator 7 & .390 & .095 & - & .429 \\
Human Annotator 8 & .389 & .090 & - & .406 \\\hline
DUC04 Participant A & \bf.382 & .071 & - & .313 \\
DUC04 Participant B & .374 & .071 & - & .310 \\
DUC04 Participant C & .374 & .073 & - & .312 \\
DUC04 Participant D & .374 & .083 & - &\bf.374 \\
DUC04 Participant E & .371 & .048 & - & .305 \\\hline
ILP      &.377            &\bf.092            &.126            &.333 \\
ILP+MC   &.342            &.072                   &.109            &.308          \\
\hline
\end{tabular}

\end{minipage}
\end{center}
\vspace{-10pt}
\caption{A comparison with official participants in DUC 2004, including 8 human annotators (1-8) and the top-5 offical participants (A-E). `-' means a metric is not available.}
\label{tab:results_ilp_duc04}
\end{table}

\subsection{Extrinsic evaluation}
\label{sec:extrinsc}

Our proposed approach is compared against a range of baselines.
They are 1) MEAD~\cite{Radev:2004}, a centroid-based summarization system that scores sentences based on length, centroid, and position;
2) LexRank~\cite{Erkan:2004}, a graph-based summarization approach based on eigenvector centrality;
3) SumBasic~\cite{Vanderwende:2007}, an approach that assumes words occurring frequently in a document cluster have a higher chance of being included in the summary;
4) Pointer-Generator Networks (PGN)~\cite{see-liu-manning:2017:Long}, a state-of-the-art neural encoder-decoder approach for abstractive summarization. The system was trained on the CNN/Daily Mail data sets~\cite{hermann2015teaching,DBLP:journals/corr/NallapatiXZ16}.
5) ILP~\cite{Kirkpatrick:2011}, a baseline ILP framework without matrix completion.

The Pointer-Generator Networks~\cite{see-liu-manning:2017:Long} describes a neural encoder-decoder architecture. 
It encourages the system to copy words from the source text via pointing, while retaining the ability to produce novel words through the generator. 
It also contains a coverage mechanism to keep track of what has been summarized, thus reducing word repetition.
The pointer-generator networks have not been tested for summarizing content contributed by multiple authors. In this study we evaluate their performance on our collection of datasets.

For the ILP-based approaches, we use bigrams as concepts (bigrams consisting of only stopwords are removed\footnote{Bigrams with one stopword are not removed because 1) they are informative (``a bike", ``the activity", ``how materials'); 2) such bigrams appear in multiple sentences and are thus helpful for matrix imputation.}) and term frequency as concept weights.
We leverage the co-occurrence statistics both within and across the entire corpus\footnote{We construct one single matrix for each entire corpus except DUC04. For example, the co-occurrence matrix for Eng includes 1492 distinct sentences and 9239 unique bigrams, from all lectures and prompts. For DUC04, we construct a matrix for each document cluster instead of the entire corpus due to its high computational cost.}. We also filtered out bigrams that appear only once in each corpus, yielding better ROUGE scores with lower computational cost. The results without using this low-frequency filtering are shown in the Appendix 
for comparison. In Table~\ref{tab:results_ilp}, we present summarization results evaluated by ROUGE~\cite{Lin:2004} and human judges.\footnote{The results on Eng are slightly different from the results published by Luo et al.~\shortcite{Luo:2016:NAACL} as we used leave-one-lecture-out cross-validation instead of 3-fold cross-validation to select the parameter $\lambda$. We also changed the order of student responses by grouping same responses together, affecting the position feature in MEAD. }

To compare with the official participants in DUC 2004~\cite{paul2004introduction}, we selected the top-5 systems submitted in the competition (ranked by R-1), together with the 8 human annotators. The results are presented in Table~\ref{tab:results_ilp_duc04}.

\vspace{0.1in}
\noindent\textbf{ROUGE.} 
It is a recall-oriented metric that compares system and reference summaries based on n-gram overlaps, which is widely used in summarization evaluation. In this work, we report ROUGE-1 (R-1), ROUGE-2 (R-2), ROUGE-SU4 (R-SU4), and ROUGE-L (R-L) scores, which respectively measure the overlap of unigrams, bigrams, skip-bigram (with a maximum gap length of 4), and longest common subsequence. 
First, there is no winner for all data sets. MEAD is the best one on camera; SumBasic is best on Stat2016 and mostly on Stat2015; ILP is best on DUC04. The ILP baseline is comparable to the best participant (Table~\ref{tab:results_ilp_duc04}) and even has the best R-2. PGN is the worst, which is not surprising since it is trained on a different data set, which may not generalize to our data sets. Our method ILP+MC is best on peer review and mostly on Eng and CS2016.
Second, compared with ILP, our method works better on Eng, CS2016, movie, and peer. 

These results show our proposed method does not always better than the ILP framework, and no single summarization system wins on all data sets. It is perhaps not surprising to some extent.
The {\bf no free lunch theorem} for machine learning~\cite{wolpert1996lack} states that, averaged overall possible data-generating distributions, every classiﬁcation algorithm has the same error rate when classifying previously unobserved points. In other words, in some sense, no machine learning algorithm is universally any better than any other~\cite{Goodfellow-et-al-2016}.

\begin{figure}[!ht]
\centering 
\includegraphics[width=\linewidth,keepaspectratio]{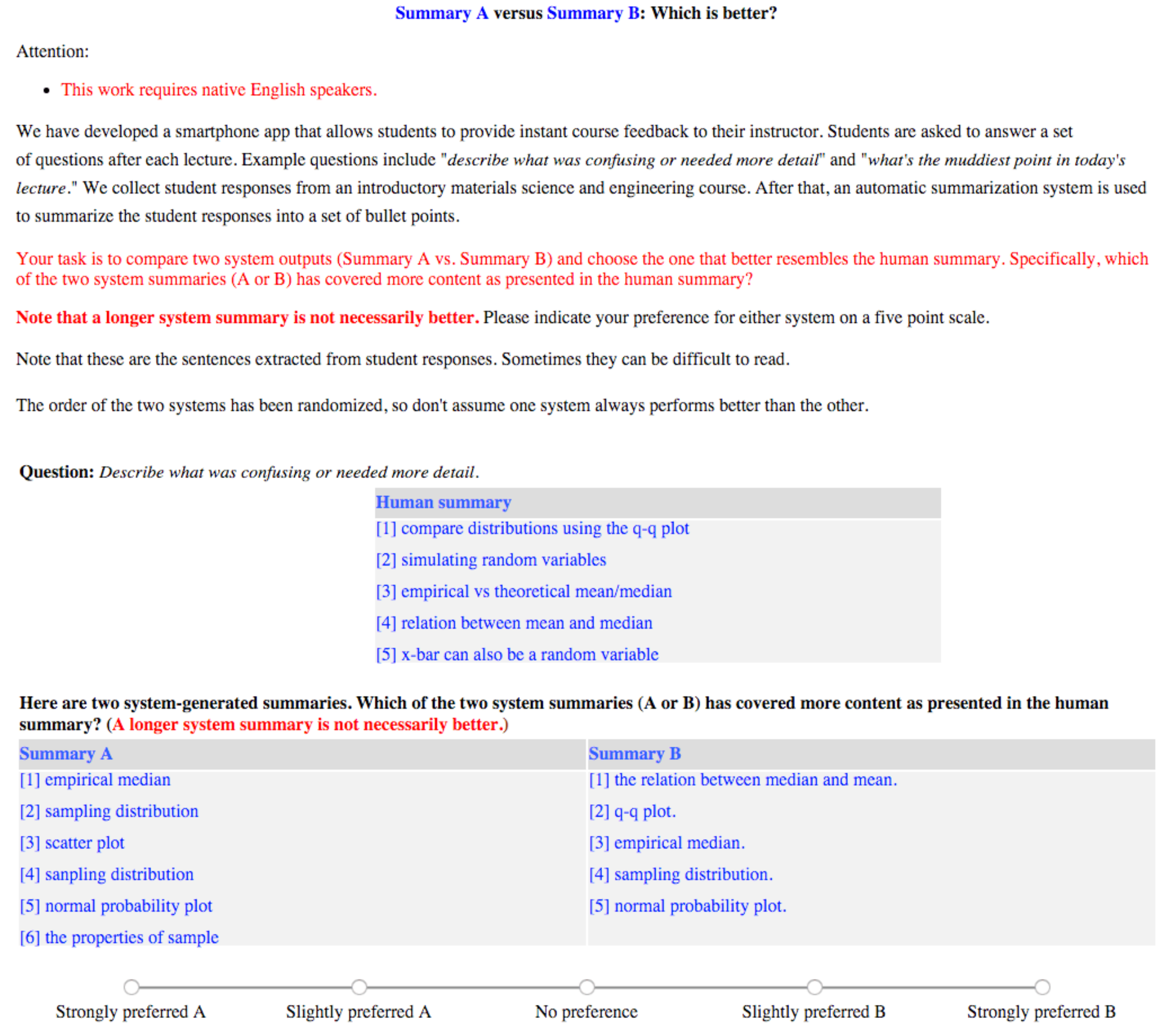}
\vspace{-20pt}
\caption{An example HIT from Stat2015, `System A' is ILP+MC and `System B' is SumBasic.}
\label{fig:hit1}
\end{figure}

\vspace{0.1in}
\noindent\textbf{Human Evaluation.} Because ROUGE cannot thoroughly capture the semantic similarity between system and reference summaries, we further perform a human evaluation.
For each task, we present a pair of system outputs in a random order, together with one human summary to five Amazon turkers. If there are multiple human summaries, we will present each human summary and the pair of system outputs to turkers. For student responses, we also present the prompt. An example Human Intelligence Task (HIT) is illustrated in Fig.~\ref{fig:hit1}.

\begin{figure}[!ht]
\centering 
\includegraphics[width=\linewidth,keepaspectratio]{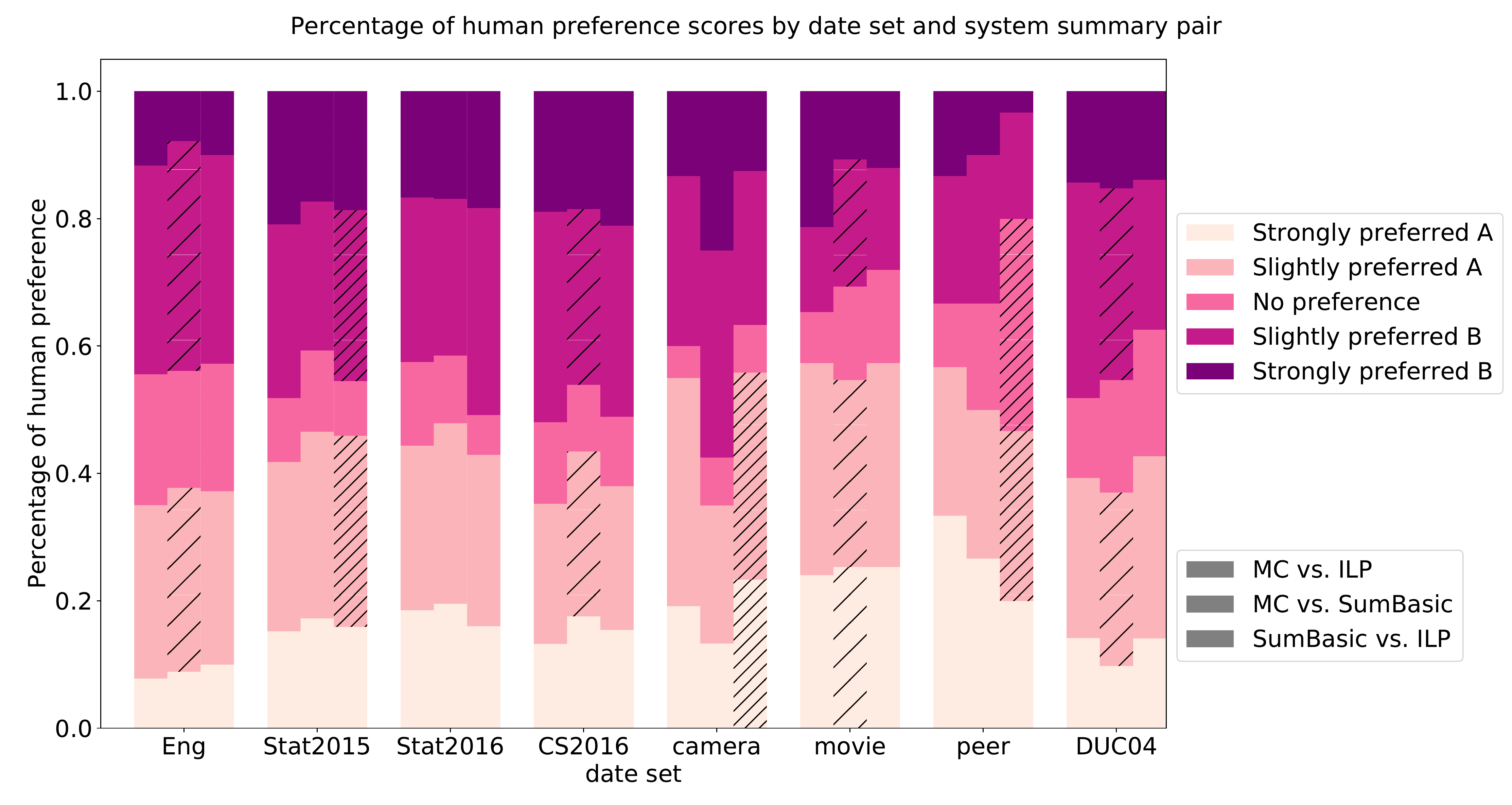}
\vspace{-20pt}
\caption{Distribution of human preference scores}
\label{fig:scores}
\end{figure}

The turkers are asked to indicate their preference for system A or B based on the semantic resemblance to the human summary on a 5-Likert scale (`Strongly preferred A', `Slightly preferred A', `No preference',  `Slightly preferred B', `Strongly preferred B').
They are rewarded \$0.04 per task. We use two strategies to control the quality of the human evaluation. First, we require the turkers to have a HIT approval rate of 90\% or above. Second, we insert some quality checkpoints by asking the turkers to compare two summaries of same text content but in different sentence orders. Turkers who did not pass these tests are filtered out.
Due to budget constraints, we conduct pairwise comparisons for three systems. 
The total number of comparisons is 3 system-system pairs $\times$ 5 turkers $\times$ (36 tasks $\times$ 1 human summaries for Eng + 44$\times$2 for Stat2015 + 48$\times$2 for Stat2016 + 46$\times$2 for CS2016 + 3$\times$8 for camera + 3$\times$5 for movie + 3$\times$2 for peer + 50 $\times$ 4 for DUC04) = 8,355. The number of tasks for each corpus is shown in Table~\ref{table:existing_dataset}. To elaborate as an example, for Stat2015, there are 22 lectures and 2 prompts for each lecture. Therefore, there are 44 tasks (22$\times$2) in total. In addition, there are 2 human summaries for each task. We selected three competitive systems (SumBasic, ILP, and ILP+MC) and therefore we have 3 system-system pairs (ILP+MC vs. ILP, ILP+MC vs. SumBasic, and ILP vs. SumBasic) for each task and each human summary. Therefore, we have 44$\times$2$\times$3=264 HITs for Stat2015. Each HIT will be done by 5 different turkers, resulting in 264$\times$5=1,320 comparisons. In total, 306 unique turkers were recruited\footnote{The turkers are anonymized.} and on average 27.3 of HITs were completed by one turker. The distribution of the human preference scores is shown in Fig.~\ref{fig:scores}.

\begin{table}[!ht]
\centering
\begin{tabular}{l|l|l|l}
\hline
  &  ILP+MC vs. ILP  &  ILP+MC vs. SumBasic  &  SumBasic vs. ILP \\
\hline\hline
Eng  &  51.1\%  &  49.4\%  &  50.9\% \\
Stat2015  &  49.9\%$^*$ (ILP+MC)  &  50.0\%  &  51.2\% \\
Stat2016  &  48.0\%  &  49.2\%  &  51.2\% \\
CS2016  &  51.3\%$^*$ (ILP+MC)  &  51.5\%  &  50.6\%$^*$ (SumBasic) \\\hline\hline
camera  &  49.2\%  &  47.5\%$^*$ (ILP+MC)  &  46.7\%$^*$ (ILP) \\
movie  &  \underline{45.3\%}  &  50.7\%$^*$ (SumBasic) &  \underline{44.0\%}$^*$ (SumBasic) \\
peer  &  53.3\%  &  \underline{43.3\%}  &  50.0\%$^*$ (ILP) \\\hline\hline
DUC04  &  48.4\%$^*$ (ILP)  &  46.4\%$^*$ (ILP+MC) &  \underline{44.0\%} \\
\hline
\end{tabular}
\vspace{-10pt}
\caption{Inter-annotator agreement measured by the percentage of individual judgements agreeing with the majority votes. $^*$ means the human preference to the two systems are significantly different and the system in parenthesis is the winner. \underline{Underline} means that it is lower than random choices (45.7\%).
}
\label{tab:results_ilp_agreement}
\end{table}

We calculate the percentage of ``wins'' (strong or slight preference) for each system among all comparisons with its counterparts.
Results are reported in the last column of Table~\ref{tab:results_ilp}\footnote{The sum of the percentage is not 100\% because there are ``no preference" choices.}.
\textsc{ILP+MC} is preferred significantly\footnote{For the significance test, we convert a preference to a score ranging from -2 to 2 (`2' means `Strongly preferred' to a system and `-2' means `Strongly preferred' to the counterpart system), and use a two-tailed paired t-test with $p < 0.05$ to compare the scores. Similar significant results can be observed if using a 3-point Likert scale  (`preferred A', `no preference', `preferred B'), except that the difference between ILP and ILP+MC is not significant for Stat2015, but significant for CS2016 and movie.} more often than ILP on Stat2015, CS2016, and DUC04. There is no significant difference between ILP+MC and SumBasic on student response data sets. Interestingly, a system with better ROUGE scores does not necessarily mean it is more preferred by humans. For example, ILP is preferred more on all three review data sets. 
Regarding the inter-annotator agreement, we find 48.5\% of the individual judgements agree with the majority votes. 
The agreement scores decomposed by data sets and system pairs are shown in Table~\ref{tab:results_ilp_agreement}. Overall, the agreement scores are pretty low, compared to an agreement score achieved by randomly clicking (45.7\%)\footnote{The random agreement score on a 5-Likert scale can be verified by a simulation experiment.}. It has several possibilities. The first one is that many turkers did click randomly (39 out of 160 failed our quality checkpoints). Unfortunately, we did not check all the turkers as we inserted the checkpoints randomly. The second possibility is that comparing two system summaries is difficult for humans, and thus it has a low agreement score. Xiong and Litman~\shortcite{xiong-litman:2014:Coling} also found that it is hard to make humans agree on the choice of summary 
sentences. A third possibility is that turkers needed to see the raw input sentences which are not shown in a HIT.

An interesting observation is that our approach produces summaries with more sentences, as shown in Table~\ref{tab:results_ilp_length}. The number of words in the summaries is approximately the same for all methods for a particular corpus, which is constrained by Eq.~\ref{eqn:ilp_length_new}. For camera, movie and peer reviews, the number of sentences in human summary is 10, and SumBasic and ILP+MC produce more sentences than ILP. It is hard for people to judge which system summaries is closer to a human summary when the summaries are long (216, 242, and 190 words for camera, movie, and peer reviews respectively). For inter-annotator agreement, 50.3\% of judgements agree with the majority votes for student response data sets, 47.6\% for reviews, and only 46.3\% for news documents. We hypothesize that for these long summaries, people may prefer short system summaries, and for short summaries, people may prefer long system summaries. We leave the examination of this finding to future work.

\begin{table}[!ht]
\centering
\begin{small}
\begin{tabular}{l|cccc|ccc|c}
\hline
&Eng&Stat2015&Stat2016&CS2016&camera&movie&peer&DUC04\\
\hline\hline
MEAD  &\ 1.6$^*$  &\ 1.3$^*$  &\ 2.2$^*$  &\ 1.1$^*$  &\ 3.0$^*$  &\ 1.7$^*$  &\ 3.3$^*$  &\ 2.5$^*$  \\
LexRank  &\ 2.8$^*$  &\ 2.4$^*$  &\ 3.0$^*$  &\ 1.9$^*$  &7.0 &\ 5.3$^*$  &\ 6.0$^*$  &\ 3.4$^*$  \\
SumBasic  &6.0 &5.6&\ 5.8$^*$  &4.2&14.7&\ 19.7$^*$  &\ 12.3$^*$  &7.7\\
ILP  &\ 4.8$^*$  &\ 3.6$^*$  &\ 3.7$^*$  &\ 2.6$^*$  &\ 14.0$^*$  &17.7&\ 12.0$^*$  &\ 5.2$^*$  \\
ILP+MC &6.4&5.6&5.3&4.3&17.3&31.3&16.7&7.5\\
\hline
\end{tabular}
\end{small}
\vspace{-10pt}
\caption{Number of sentences in the output summaries. $^*$ means it is significantly different to ILP+MC (p $<$ 0.05) using a two-tailed paired t-test.
}
\label{tab:results_ilp_length}
\end{table}

Table~\ref{tab:output} presents example system outputs. This
offers an intuitive understanding of our proposed approach.

\begin{table}[!ht]
\centering
\renewcommand{\arraystretch}{1.0}
\begin{tabular}{ l }
\hline

\rule{0pt}{3ex}\textbf{Prompt}\\
\textit{Describe what you found most interesting in today's class}\\

\rule{0pt}{3ex}{\bf Reference Summary}\\
- unit cell direction drawing and indexing\\
- real world examples\\
- importance of cell direction on materials properties\\

\rule{0pt}{3ex}{\bf System Summary (\textsc{ILP Baseline})}\\
- drawing and indexing unit cell direction\\
- it was interesting to understand how to find apf and fd from last weeks class\\
\rule[-2.2ex]{0pt}{0pt}- south pole explorers died due to properties of tin\\

\rule{0pt}{3ex}{\bf System Summary (\textsc{ILP+MC})}\\
- crystal structure directions\\
- surprisingly i found nothing interesting today .\\
- unit cell indexing\\
- vectors in unit cells\\
- unit cell drawing and indexing\\
\rule[-2.2ex]{0pt}{0pt}- the importance of cell direction on material properties\\
\hline
\end{tabular}
\caption{Example reference and system summaries. }
\label{tab:output}
\end{table}

\section{Analysis of Influential Factors}
\label{chapter:sec:factors}

In this section, we want to investigate the impact of the low-rank approximation process to the ILP framework. Therefore, in the following experiments, we focus on the direct comparison with the ILP and ILP+MC and leave the comparison to other baselines as future work.
The proposed method achieved better summarization performance on Eng, CS2016, movie, and peer than the ILP baseline. Unfortunately, it does not work as expected on two courses for student responses (Stat2015 and Stat2016), review camera and news documents. This leaves the research question when and why the proposed method works better. In order to investigate what are key factors that impact the performance, we would like to perform additional experiments using synthesized data sets.

\begin{table}[!ht]
\centering
\begin{small}
\begin{tabular}{l|l|l}
\hline
 & \textbf{id} & description \\\hline
\textbf{Input} & 1 & \tabitem {\bf genre}: belonging to student response/review/news \\
& 2 & \tabitem {\bf T}: number of \underline{t}asks\\ 
& 3 & \tabitem {\bf au}: number of \underline{au}thors\\ 
& 4 & \tabitem {\bf M*N}: size of $A$\\
& 5 & \quad \tabitem {\bf M}: number of sentences in total\\
& 6 & \quad \tabitem {\bf N}: number of bigrams in total\\
& 7 & \quad \tabitem {\bf M/T}: number of sentences per task \\
& 8 & \quad \tabitem {\bf N/T}: number of bigrams per task \\
& 9 & \quad \tabitem {\bf N/M}: number of bigrams per sentence \\
& 10 & \quad \tabitem {\bf W/T}: number of \underline{w}ords per \underline{t}ask \\
& 11 & \quad \tabitem {\bf W/M}: number of \underline{w}ords per sentence \\
& 12 & \tabitem {\bf s}: \underline{s}parsity ratio, ratio of 0 cells in $A$ per task \\
& 13 & \tabitem {\bf b=1}: ratio of bigrams appear only once\\
& 14 & \tabitem {\bf b$>$1}: ratio of bigrams appear more than once\\
& 15 & \tabitem {\bf H}: Shannon's diversity index, defined as $H = -\sum_i p_i\ln p_i$, \\
&  & \quad \quad where $p_i$ is the frequency of bigram $i$ divided by \\
&  & \quad \quad total number of bigrams in a task\\
\hline
\textbf{Summaries} & 16 & \tabitem {\bf L}: \underline{l}ength of human summaries in number of words\\
& 17 & \tabitem {\bf hs}: number of \underline{h}uman \underline{s}ummaries per task\\

& 18 & \tabitem {\bf r}: compression \underline{r}atio, length of human summaries compared to \\
&  & \quad \quad length of input documents \\
& 19 & \tabitem {\bf $\alpha_{b>0}$}: \underline{a}bstraction ratio, how many of bigrams in {\bf human } \\
&  & \quad \quad {\bf summaries} appeared in the original documents at least once \\ 
& 20 & \quad\tabitem {\bf $\alpha_{b=0}$}: ratio of bigrams in human summaries that are \\
&  & \quad \quad not in the input \\
& 21 & \quad\tabitem {\bf $\alpha_{b=1}$}: ratio of bigrams in human summaries that are \\
&  & \quad \quad in the input only once\\
& 22 & \quad\tabitem {\bf $\alpha_{b>1}$}: ratio of bigrams in human summaries that are \\
&  & \quad \quad in the input more than once\\
& 23 & \tabitem {\bf $\beta_{b=1}$}: ratio of bigrams in the {\bf input} appear only {\bf once} \\
&  & \quad \quad but selected by human(s)\\
& 24 & \quad\tabitem {\bf $\beta_{b=2}$}: ratio of bigrams in the input appear {\bf twice} and \\
&  & \quad \quad selected by human(s)\\
& 25 & \quad\tabitem {\bf $\beta_{b=3}$}: ratio of bigrams in the input appear {\bf three times}  \\
&  & \quad \quad and selected by human(s)\\
& 26 & \quad\tabitem {\bf $\beta_{b=4}$}: ratio of bigrams in the input appear {\bf four times} \\
&  & \quad \quad and selected by human(s)\\
& 27 & \quad\tabitem {\bf $\beta_{b>1}$}: ratio of bigrams in the input appear {\bf more than once} \\
&  & \quad \quad and selected by human(s)\\
\hline
\end{tabular}

\end{small}
\caption{Attributes description, extracted from the input and the human reference summaries.}
\label{tab:feature_descriptor}
\end{table}

\begin{table}[!ht]
\begin{center}
\begin{scriptsize}
\begin{tabular}{l| l| cccc| ccc| c}
\hline
id  &  name  &  Eng  &  Stat2015  &  Stat2016  &  CS2016  &  camera  &  movie  &  peer  &  DUC04 \\\hline\hline
1  &  genre  &  response  &  response  &  response  &  response  &  review  &  review  &  review  &  news \\
2  &  T  &  36  &  44  &  48  &  46  &  3  &  3  &  3  &  50 \\
3  &  au  &  37.7  &  39.3  &  42.2  &  22.4  &  18.0  &  18  &  18  &  10 \\
4  &  M*N  &  13.8 & 10.8 & 7.2 & 7.4  &  0.9 & 15.7 & 0.7  &  2291.7 \\
5  &  M  &  1492  &  1696  &  1660  &  1162  &  255  &  985  &  241  &  11566 \\
6  &  N  &  9239  &  6366  &  4329  &  6409  &  3716  &  15934  &  2934  &  198140 \\
7  &  M/T  &  41.4  &  38.5  &  34.6  &  25.3  &  85.0  &  328.3  &  80.3  &  231.3 \\
8  &  N/T  &  256.6  &  144.7  &  90.2  &  139.3  &  1238.7  &  5311.3  &  978.0  &  3962.8 \\
9  &  N/M  &  6.2  &  3.8  &  2.6  &  5.5  &  14.6  &  16.2  &  12.2  &  17.1 \\
10  &  W/T  &  375.4  &  233.1  &  149.3  &  223.1  &  1927.0  &  8014.0  &  1543.7  &  5171.6 \\
11  &  W/M  &  9.1  &  6.0  &  4.3  &  8.8  &  22.7  &  24.4  &  19.2  &  22.4 \\
12  &  s  &  97.2\%  &  96.6\%  &  96.0\%  &  95.4\%  &  98.5\%  &  99.6\%  &  98.5\%  &  99.4\% \\
13  &  $b=1$  &  90.3\%  &  90.1\%  &  87.6\%  &  94.0\%  &  94.7\%  &  92.6\%  &  91.1\%  &  85.5\% \\
14  &  $b>1$  &  9.7\%  &  9.9\%  &  12.4\%  &  6.0\%  &  5.3\%  &  7.4\%  &  8.9\%  &  14.5\% \\
15  &  $H$  &  5.282  &  4.590  &  4.007  &  4.703  &  6.894  &  8.314  &  6.617  &  7.844 \\\hline
16  &  L  &  30  &  15  &  13  &  16  &  216  &  242  &  190  &  105 \\
17  &  hs  &  1  &  2  &  2  &  2  &  8  &  5  &  2  &  4 \\
18  &  r  &  8.8\%  &  7.6\%  &  10.9\%  &  8.3\%  &  13.1\%  &  3.1\%  &  13.5\%  &  2.4\% \\
19  &  $\alpha_{b>0}$  &  48.8\%  &  46.5\%  &  56.4\%  &  45.8\%  &  96.7\%  &  97.6\%  &  95.9\%  &  37.0\% \\
20  &  $\alpha_{b=0}$  &  51.2\%  &  53.5\%  &  43.6\%  &  54.2\%  &  3.3\%  &  2.4\%  &  4.1\%  &  63.0\% \\
21  &  $\alpha_{b=1}$  &  34.1\%  &  18.1\%  &  20.9\%  &  25.6\%  &  84.9\%  &  76.4\%  &  77.1\%  &  15.9\% \\
22  &  $\alpha_{b>1}$  &  14.7\%  &  28.4\%  &  35.5\%  &  20.2\%  &  11.8\%  &  21.2\%  &  18.8\%  &  21.1\% \\
23  &  $\beta_{b=1}$  &  3.3\%  &  2.7\%  &  4.3\%  &  3.7\%  &  45.8\%  &  11.2\%  &  23.3\%  &  1.7\% \\
24  &  $\beta_{b=2}$  &  8.5\%  &  16.5\%  &  28.2\%  &  25.1\%  &  65.3\%  &  20.4\%  &  40.3\%  &  7.2\% \\
25  &  $\beta_{b=3}$  &  12.5\%  &  39.0\%  &  58.8\%  &  57.4\%  &  79.3\%  &  31.8\%  &  53.8\%  &  13.7\% \\
26  &  $\beta_{b=4}$  &  33.3\%  &  61.1\%  &  76.9\%  &  50.0\%  &  90.9\%  &  42.9\%  &  50.0\%  &  22.1\% \\
27  &  $\beta_{b>1}$  &  12.3\%  &  28.0\%  &  45.2\%  &  37.0\%  &  70.0\%  &  27.7\%  &  46.0\%  &  12.0\% \\
\hline
\end{tabular}
\end{scriptsize}
\end{center}
\caption{Attributes extracted from the input and the human reference summaries. The numbers in the row of $M*N$ are divided by $10^6$. The description of each attribute is shown in Table~\ref{tab:feature_descriptor}.}
\label{tab:feature}
\end{table}

A variety of attributes that might impact the performance are summarized in Table~\ref{tab:feature_descriptor}, categorized into two types. The {\bf input} attributes are extracted from the input original documents and the {\bf summaries} attributes are extracted from human summaries and the input documents as well. Here are some important attributes we expect to have a big impact on the performance. 

\begin{itemize}
\item $M*N$ is the size of the summarization task, represented by the size of the co-occurrence matrix $A$, as shown in Eq.~\ref{eqn:orig_ilp_bigram} and Eq.~\ref{eqn:orig_ilp_sentence}. Generally, the bigger the matrix, the more difficult it is to find an optimal solution of low-rank approximation as there are more parameters. Note, $A$ is an $N \times M$ matrix, where $N$ is the number of unique concepts, and $M$ is the number of sentences. 

\item For the sparsity ratio $s$, if the matrix is too sparse, there will not be enough information within $A$ to have a good estimate of the completed matrix after imputation. In contrast, if the matrix is not sparse at all (e.g., all authors use the same term for a concept), there will be no benefit to performing low-rank approximation. 

\item The Shannon's diversity index $H$ measures the degree of bigram diversity. The more diverse the bigram distribution, the smaller the corresponding Shannon entropy.

\item The abstraction ratios $\alpha_{b=0}$, $\alpha_{b=1}$, $\alpha_{b>1}$ capture in what degree annotators use words from the input or use their own.  

\item $\beta_{b=1}$, $\beta_{b=2}$, $\beta_{b=3}$, $\beta_{b=4}$, $\beta_{b>1}$ intend to capture how humans create the summaries in terms of whether more frequent bigrams are more likely to be selected by humans.
\end{itemize}

The attributes extracted from the corpora are shown in Table~\ref{tab:feature}.  Note, a bigram that appears more often in original documents has a better chance to be included in human summaries as indicated by $\beta_{b=1}$, $\beta_{b=2}$, $\beta_{b=3}$, and $\beta_{b=4}$. This verifies our choice to cut low-frequency bigrams.

According to the ROUGE scores, our method works better on Eng, CS2016, movie, and peer (Table~\ref{tab:results_ilp}). If we group each attribute into two groups, corresponding to whether ILP+MC works better, we do not find significant differences among these attributes. To further understand which factors impact the performance and have more predictive power, we train a binary classification decision tree by treating the 4 working corpora as positive examples and the remaining 4 as negative examples.

According to the decision tree model, there is only one decision point in the tree: $\alpha_{b=1}$, the ratio of bigrams in human summaries that are in the input only once. Generally, our proposed method works if $\alpha_{b=1} > 23.2\%$, except for camera. When $\alpha_{b=1}$ is low, it means that annotators either adopt concepts that appear multiple times or just use their own. In this case, the frequency-based weighting (i.e., $w_i$ in Eq.~\ref{eqn:orig_ilp_obj}) can capture the concepts that appear multiple times. On the other hand, when $\alpha_{b=1}$ is high, it means that a big number of bigrams appeared only once in the input document. In this case, annotators have difficulty selecting a representative one due to the ambiguous choice. Therefore, we hypothesize,

\begin{itemize}
\item {\bf H2}: The ILP framework benefits more from low-rank approximation when $\alpha_{b=1}$ is higher.
\end{itemize}

To test the predictive power of this attribute, we want to test it on new data sets. Unfortunately, creating new data sets with gold-standard human summaries is expensive and time-consuming, and the new data set may not have the desired property within a certain range of $\alpha_{b=1}$. Therefore, we propose to manipulate the ratio and create new data sets using the existing data sets without additional human annotation. $\alpha_{b=1}$ can be represented as follows,

\begin{align}
\label{eqn:ratio}
\displaystyle 
\alpha_{b=1} = \frac{\sum_i \sigma_i \cdot \phi_{w_i=1}}{\sum_i \sigma_i}
\end{align}

where 
$$\sigma_i = \begin{cases}
1 & \text{if bigram $i$ appears in the human summary}\\
0 & else
\end{cases}$$

$$\phi_{w_i=1} = \begin{cases}
1 & \text{if } w_i=1, w_i\text{ is the weight of the bigram }$i$\\
0 & else
\end{cases}$$

There are two different ways to control the ratio, both involving removing input sentences with certain constraints.

\begin{itemize}
\item To increase this ratio, we remove sentences with bigrams that appear multiple times so that there will be more bigrams that appear once (i.e., increase $\sigma_i \cdot \phi_{w_i=1}$) and thus increase the numerator. For example, if a bigram in a human summary appears in two input sentences (e.g., S1 and S2), we can randomly remove one of them (either S1 or S2) to make the bigram appear only once in the input. 
Note that we keep sentences that have bigrams appearing multiple times and a bigram appearing only once as well, so that we guarantee that all the input sentences with a unique bigram in human summaries are kept and removing other sentences can only increase the ratio.

\item To decrease this ratio, we remove the sentences with bigrams that appear only once in order to decrease the numerator. This will reduce the bigram frequency $w_i$ from 1 to 0. Similarly, we keep sentences that contain bigrams appearing multiple times so that removing sentences will not increase the ratio.
\end{itemize}

In this way, we obtained different levels of $\alpha_{b=1}$ by deleting sentences. The ROUGE scores on the synthesized corpus are shown in Table~\ref{tab:synthesize_res}. 

\begin{table}
\centering
\begin{minipage}{0.8\textwidth}
\begin{footnotesize}
\begin{tabular}{l | l | l | l| l | l|l}
\hline\hline 
 & $\alpha_{b=1}$ & \multicolumn{1}{l|}{\textbf{System}} & \textbf{R-1} & \textbf{R-2} & \textbf{R-SU4} & \textbf{R-L} \\
 \hline
		&26.5	&ILP		&.341		&\bf.112	&.121		&.329 \\
		&		&ILP+MC		&\bf.378$^+$&\bf.112	&\bf.137	&\bf{.366}$^+$\\
Eng		&34.1	&ILP		&.364		&.123		&.135		&.347\\
		&		&ILP+MC		&\bf.392$^+$&\bf.130	&\bf.150	&\bf{.380}$^+$\\
		&36.0	&ILP		&.358		&.119		&.124		&.345\\
		&		&ILP+MC		&\bf.397$^+$&\bf.126	&\bf.155	&\bf{.379}\\\hline
		&11.9	&ILP		&.401		&.183		&.160		&.393\\
		&		&ILP+MC		&.379		&.167		&.158		&.369\\
Stat2015&18.1	&ILP		&.405		&.186		&.165		&.397\\
		&		&ILP+MC		&.401		&.183		&\bf.173	&.391\\
		&21.0	&ILP		&.394		&.172		&.160		&.384\\
		&		&ILP+MC		&.350$^-$	&.144$^-$	&.140		&.341$^-$\\\hline
		&13.2	&ILP		&.467		&.252		&.202		&.454\\
		&		&ILP+MC		&.461		&.209$^-$	&.200		&.443\\
Stat2016&20.9	&ILP		&.482		&.262		&.215		&.468\\
		&		&ILP+MC		&.457		&.214$^-$	&.198		&.441\\
		&23.7	&ILP		&.455		&.244		&.198		&.441\\
		&		&ILP+MC		&\bf.476	&.210$^-$	&\bf.203	&\bf{.460}\\\hline
		&11.0	&ILP		&.362		&.138		&.134		&.346\\
		&		&ILP+MC		&\bf.376	&.135		&\bf.151	&\bf{.360}\\
CS2016	&25.6	&ILP		&.374		&.141		&.137		&.356\\
		&		&ILP+MC		&\bf.398	&\bf.154	&\bf.158$^+$&\bf{.383}$^+$\\
		&34.2	&ILP		&.296		&.091		&.087		&.281\\
		&		&ILP+MC		&\bf.318	&.085		&\bf.100	&\bf{.306}\\\hline
		&78.7	&ILP		&.453		&.166		&.182		&.426\\
		&		&ILP+MC		&.437		&.147		&.172		&.407\\
camera	&84.9	&ILP		&.457		&.165		&.181		&.427\\
		&		&ILP+MC		&.447		&.157		&.176		&.418\\
		&85.8	&ILP		&.452		&.156		&.178		&.422\\
		&		&ILP+MC		&\bf.454	&\bf.159	&\bf.183	&\bf{.427}\\\hline
		&71.9	&ILP		&.439		&.107		&.172		&.397\\
		&		&ILP+MC		&.417		&.103		&.159		&.392\\
movie	&76.4	&ILP		&.435		&.091		&.167		&.397\\
		&		&ILP+MC		&\bf.436	&\bf.106$^+$&\bf.169	&\bf{.409}\\
		&76.8	&ILP		&.435		&.109		&.170		&.387\\
		&		&ILP+MC		&.411		&.102		&.155		&.383\\\hline
		&71.3	&ILP		&.467		&.206		&.188		&.448\\
		&		&ILP+MC		&.442		&.193		&.158		&.421\\
peer	&77.1	&ILP		&.466		&.199		&.183		&.445\\
		&		&ILP+MC		&\bf.491	&\bf.261	&\bf.195	&\bf{.469}\\
		&78.7	&ILP		&.488		&.242		&.199		&.475\\
		&		&ILP+MC		&.447		&.183		&.162		&.425\\\hline
		&13.9	&ILP		&.376		&.092		&.124		&.332\\
		&		&ILP+MC		&.349$^-$		&.074$^-$		&.113$^-$		&.314$^-$\\
DUC04	&15.9	&ILP		&.377		&.092		&.126		&.333\\
		&		&ILP+MC		&.342$^-$		&.072$^-$		&.109$^-$		&.308$^-$\\
		&16.5	&ILP		&.375		&.093		&.123		&.332\\
		&		&ILP+MC		&.349$^-$		&.074$^-$		&.113$^-$		&.314$^-$\\
  \hline
  \hline
\end{tabular}
\end{footnotesize}

\end{minipage}
\vspace{-10pt}
\caption{ROUGE scores on synthesized corpora. {\bf Bold} scores indicate our approach ILP+MC is better than ILP. $^+$ and $^-$ mean a score is significantly better and worse respectively ($p < 0.05$) using a two-tailed paired t-test.}
\label{tab:synthesize_res}
\end{table}

Our hypothesis {\bf H2} is partially valid. When increasing the ratio, ILP+MC has a relative advantage gain over ILP. For example, for Stat2015, ILP+MC is not significantly worse than ILP any more when increasing the ratio from 11.9 to 18.1. For camera, ILP+MC becomes better than ILP when increasing the ratio from 84.9 to 85.8.
For Stat2016, CS2016, Eng, more improvements or significant improvements can be found for ILP+MC compared to ILP when increasing the ratio. However, for movie and peer review, ILP+MC is worse than ILP when increasing the ratio.  

We have investigated a number of attributes that might impact the performance of our proposed method. Unfortunately, we do not have a conclusive answer when our method works better. However, we would like to share some thoughts about it.

First, our proposed method works better on two student responses courses (Eng and CS2016), but not the other two (Stat2015 and Stat2016). An important factor we ignored is that the students from the other two courses are not native English speakers, resulting in significantly shorter responses (4.3 $<$ 6.0 $<$ 8.8, 9.1, $p < 0.01$, Table~\ref{tab:feature}, the row with id=11). With shorter sentences, there will be less context to leverage the low-rank approximation.

Second, our proposed method works better on movie and peer reviews, but not camera reviews. As pointed out by Xiong~\shortcite{xiong2015helpfulness}, both movie reviews and peer reviews are potentially more complicated
than the camera reviews, as the review content consists of both the reviewer's evaluations
of the subject (e.g., a movie or paper) and the reviewer's references of the subject, where
the subject itself is full of content (e.g., movie plot, papers). In contrast, such references
in product reviews are usually the mentions of product components or properties, which
have limited variations. 
This characteristic makes review summarization more challenging in these two domains. 

\section{Conclusion}
\label{sec:conclusion}

We made the first effort to summarize student feedback using an Integer Linear Programming framework with a low-rank matrix approximation, and applied it to different types of data sets including news articles, product, and peer reviews.
Our approach allows sentences to share co-occurrence statistics and alleviates sparsity issue.
Our experiments showed that the proposed approach performs better against a range of baselines on the student response Eng and CS2016 on ROUGE scores, but not other courses. 

ROUGE is often adopted in research papers to evaluate the quality of summarization because it is fast and is correlated well to human evaluation~\cite{Lin:2004,graham:2015:EMNLP}. However, it is also criticized that ROUGE cannot thoroughly capture the semantic similarity between system and reference summaries. Different alternatives have been proposed to enhance ROUGE. For example, Graham~\shortcite{rankel2016statistical} proposed to use content-oriented features in conjunction with linguistic features. Similarly, Cohan and Goharian~\shortcite{COHAN16.1144} proposed to use content relevance. At the same time, many researchers supplement ROUGE with a manual evaluation. This is why we conduct evaluations using both ROUGE and human evaluation in this work. 

However, we found that a system with better ROUGE scores does not necessarily mean it is more preferred by humans (\S\ref{sec:extrinsc}). For example, ILP is preferred more on all three review data sets even if it got lower ROUGE scores than the other systems. It coincides with the fact that the ILP generated shorter summaries in terms of the number of sentences than the other two systems (Table~\ref{tab:results_ilp_length}).

We also investigated a variety of attributes that might impact the performance on a range of data sets. Unfortunately, we did not have a conclusive answer when our method will work better.

In the future, we would like to conduct a large-scale intrinsic evaluation to examine whether the low-rank matrix approximation captures similar bigrams or not and want to investigate more attributes, such as new metrics for diversity. We would like to explore the opportunities by combing a vector sentence representation learned by a neural network and the ILP framework.

\bibliographystyle{nle}
\bibliography{ref_fei_new,references,ref_self,ref_summarization,ref_clustering,ref_fei}

\begin{thebibliography}{}

\bibitem[\protect\citename{Almeida and Martins, }2013]{Almeida:2013}
Almeida, M. and Martins, A. (2013).
\newblock Fast and robust compressive summarization with dual decomposition and
  multi-task learning.
\newblock In {\em Proceedings of the 51st Annual Meeting of the Association for
  Computational Linguistics (Volume 1: Long Papers)}, pages 196--206, Sofia,
  Bulgaria. Association for Computational Linguistics.

\bibitem[\protect\citename{Barzilay et~al., }1999]{Barzilay:1999}
Barzilay, R., McKeown, K.~R., and Elhadad, M. (1999).
\newblock Information fusion in the context of multi-document summarization.
\newblock In {\em Proceedings of the Annual Meeting of the Association for
  Computational Linguistics (ACL)}.

\bibitem[\protect\citename{Berg-Kirkpatrick et~al., }2011]{Kirkpatrick:2011}
Berg-Kirkpatrick, T., Gillick, D., and Klein, D. (2011).
\newblock Jointly learning to extract and compress.
\newblock In {\em Proceedings of the Annual Meeting of the Association for
  Computational Linguistics (ACL)}.

\bibitem[\protect\citename{Boud et~al., }2013]{boud:2013}
Boud, D., Keogh, R., Walker, D., et~al. (2013).
\newblock {\em Reflection: Turning experience into learning}.
\newblock Routledge.

\bibitem[\protect\citename{Boudin et~al., }2015]{boudin2015concept}
Boudin, F., Mougard, H., and Favre, B. (2015).
\newblock Concept-based summarization using integer linear programming: From
  concept pruning to multiple optimal solutions.
\newblock In {\em Proceedings of the 2015 conference on empirical methods in
  natural language processing}, pages 1914--1918.

\bibitem[\protect\citename{Cao et~al., }2018]{Cao:2018}
Cao, Z., Wei, F., Li, W., and Li, S. (2018).
\newblock Faithful to the original: {F}act aware neural abstractive
  summarization.
\newblock In {\em Proceedings of the AAAI Conference on Artificial Intelligence
  (AAAI)}.

\bibitem[\protect\citename{Carbonell and Goldstein, }1998]{Carbonell:1998}
Carbonell, J. and Goldstein, J. (1998).
\newblock The use of {MMR}, diversity-based reranking for reordering documents
  and producing summaries.
\newblock In {\em Proceedings of the 21st Annual International ACM SIGIR
  Conference on Research and Development in Information Retrieval}, SIGIR '98,
  pages 335--336, New York, NY, USA. ACM.

\bibitem[\protect\citename{Celikyilmaz et~al.,
  }2018]{celikyilmaz-EtAl:2018:N18-1}
Celikyilmaz, A., Bosselut, A., He, X., and Choi, Y. (2018).
\newblock Deep communicating agents for abstractive summarization.
\newblock In {\em Proceedings of the 2018 Conference of the North American
  Chapter of the Association for Computational Linguistics: Human Language
  Technologies, Volume 1 (Long Papers)}, pages 1662--1675, New Orleans,
  Louisiana. Association for Computational Linguistics.

\bibitem[\protect\citename{Chen et~al., }2016]{Chen:2016}
Chen, Q., Zhu, X., Ling, Z.-H., Wei, S., and Jiang, H. (2016).
\newblock Distraction-based neural networks for document summarization.
\newblock In {\em Proceedings of the Twenty-Fifth International Joint
  Conference on Artificial Intelligence (IJCAI)}.

\bibitem[\protect\citename{Cho, }2008]{cho2008machine}
Cho, K. (2008).
\newblock Machine classification of peer comments in physics.
\newblock In {\em Educational Data Mining 2008}, pages 192--196.

\bibitem[\protect\citename{Cohan et~al., }2018]{N18-2097-Cohan-2018}
Cohan, A., Dernoncourt, F., Kim, D.~S., Bui, T., Kim, S., Chang, W., and
  Goharian, N. (2018).
\newblock A discourse-aware attention model for abstractive summarization of
  long documents.
\newblock In {\em Proceedings of the 2018 Conference of the North American
  Chapter of the Association for Computational Linguistics: Human Language
  Technologies, Volume 2 (Short Papers)}, pages 615--621. Association for
  Computational Linguistics.

\bibitem[\protect\citename{Cohan and Goharian,
  }2015]{cohan-goharian:2015:EMNLP}
Cohan, A. and Goharian, N. (2015).
\newblock Scientific article summarization using citation-context and article's
  discourse structure.
\newblock In {\em Proceedings of the 2015 Conference on Empirical Methods in
  Natural Language Processing}, pages 390--400, Lisbon, Portugal. Association
  for Computational Linguistics.

\bibitem[\protect\citename{Cohan and Goharian, }2016]{COHAN16.1144}
Cohan, A. and Goharian, N. (2016).
\newblock Revisiting summarization evaluation for scientific articles.
\newblock In Chair), N. C.~C., Choukri, K., Declerck, T., Goggi, S., Grobelnik,
  M., Maegaard, B., Mariani, J., Mazo, H., Moreno, A., Odijk, J., and
  Piperidis, S., editors, {\em Proceedings of the Tenth International
  Conference on Language Resources and Evaluation (LREC 2016)}, Paris, France.
  European Language Resources Association (ELRA).

\bibitem[\protect\citename{Cohan and Goharian, }2017]{cohan2017scientific}
Cohan, A. and Goharian, N. (2017).
\newblock Scientific document summarization via citation contextualization and
  scientific discourse.
\newblock {\em International Journal on Digital Libraries}, pages 1--17.

\bibitem[\protect\citename{Conroy and Davis, }2015]{conroy-davis:2015:VSM-NLP}
Conroy, J. and Davis, S. (2015).
\newblock Vector space models for scientific document summarization.
\newblock In {\em Proceedings of the 1st Workshop on Vector Space Modeling for
  Natural Language Processing}, pages 186--191, Denver, Colorado. Association
  for Computational Linguistics.

\bibitem[\protect\citename{Conroy et~al., }2013]{conroy-EtAl:2013:MultiLing}
Conroy, J., Davis, S.~T., Kubina, J., Liu, Y.-K., O'Leary, D.~P., and
  Schlesinger, J.~D. (2013).
\newblock Multilingual summarization: Dimensionality reduction and a step
  towards optimal term coverage.
\newblock In {\em Proceedings of the MultiLing 2013 Workshop on Multilingual
  Multi-document Summarization}, pages 55--63, Sofia, Bulgaria. Association for
  Computational Linguistics.

\bibitem[\protect\citename{Dang and Owczarzak, }2008]{Dang:2008}
Dang, H.~T. and Owczarzak, K. (2008).
\newblock Overview of the {TAC} 2008 update summarization task.
\newblock In {\em Proceedings of Text Analysis Conference (TAC)}.

\bibitem[\protect\citename{Durrett et~al., }2016]{Durrett:2016:ACL}
Durrett, G., Berg-Kirkpatrick, T., and Klein, D. (2016).
\newblock Learning-based single-document summarization with compression and
  anaphoricity constraints.
\newblock In {\em Proceedings of the 54th Annual Meeting of the Association for
  Computational Linguistics (Volume 1: Long Papers)}, pages 1998--2008, Berlin,
  Germany. Association for Computational Linguistics.

\bibitem[\protect\citename{Erkan and Radev, }2004]{Erkan:2004}
Erkan, G. and Radev, D.~R. (2004).
\newblock Lex{R}ank: Graph-based lexical centrality as salience in text
  summarization.
\newblock {\em Journal of Artificial Intelligence Research}, 22(1):457--479.

\bibitem[\protect\citename{Fan et~al., }2015]{Fan:2015}
Fan, X., Luo, W., Menekse, M., Litman, D., and Wang, J. (2015).
\newblock {CourseMIRROR}: Enhancing large classroom instructor-student
  interactions via mobile interfaces and natural language processing.
\newblock In {\em Works-In-Progress of ACM Conference on Human Factors in
  Computing Systems}. ACM.

\bibitem[\protect\citename{Fan et~al., }2017]{Fan:2017:IUI}
Fan, X., Luo, W., Menekse, M., Litman, D., and Wang, J. (2017).
\newblock Scaling reflection prompts in large classrooms via mobile interfaces
  and natural language processing.
\newblock In {\em Proceedings of 22nd ACM Conference on Intelligent User
  Interfaces (IUI 2017)}.

\bibitem[\protect\citename{Galanis et~al., }2012]{Galanis:2012}
Galanis, D., Lampouras, G., and Androutsopoulos, I. (2012).
\newblock Extractive multi-document summarization with integer linear
  programming and support vector regression.
\newblock In {\em Proceedings of COLING}.

\bibitem[\protect\citename{Gerani et~al., }2014]{Gerani:2014}
Gerani, S., Mehdad, Y., Carenini, G., Ng, R.~T., and Nejat, B. (2014).
\newblock Abstractive summarization of product reviews using discourse
  structure.
\newblock In {\em Proceedings of the Conference on Empirical Methods in Natural
  Language Processing (EMNLP)}.

\bibitem[\protect\citename{Gillick and Favre, }2009]{Gillick:2009:NAACL}
Gillick, D. and Favre, B. (2009).
\newblock A scalable global model for summarization.
\newblock In {\em Proceedings of the Workshop on Integer Linear Programming for
  Natural Langauge Processing}, pages 10--18. Association for Computational
  Linguistics.

\bibitem[\protect\citename{Gillick et~al., }2008]{Gillick:2008}
Gillick, D., Favre, B., and Hakkani-T{\"u}r, D. (2008).
\newblock The {ICSI} summarization system at {TAC} 2008.
\newblock In {\em Proceedings of TAC}.

\bibitem[\protect\citename{Gkatzia et~al., }2013]{gkatzia-EtAl:2013:ENLG}
Gkatzia, D., Hastie, H., Janarthanam, S., and Lemon, O. (2013).
\newblock Generating student feedback from time-series data using reinforcement
  learning.
\newblock In {\em Proceedings of ENLG}.

\bibitem[\protect\citename{Goldberg and Levy, }2014]{goldberg2014word2vec}
Goldberg, Y. and Levy, O. (2014).
\newblock word2vec explained: Deriving {M}ikolov et al.'s negative-sampling
  word-embedding method.
\newblock {\em arXiv preprint arXiv:1402.3722}.

\bibitem[\protect\citename{Goodfellow et~al., }2016]{Goodfellow-et-al-2016}
Goodfellow, I., Bengio, Y., and Courville, A. (2016).
\newblock {\em Deep Learning}.
\newblock MIT Press.
\newblock \url{http://www.deeplearningbook.org}.

\bibitem[\protect\citename{Graham, }2015]{graham:2015:EMNLP}
Graham, Y. (2015).
\newblock Re-evaluating automatic summarization with bleu and 192 shades of
  rouge.
\newblock In {\em Proceedings of the 2015 Conference on Empirical Methods in
  Natural Language Processing}, pages 128--137, Lisbon, Portugal. Association
  for Computational Linguistics.

\bibitem[\protect\citename{Grusky et~al., }2018]{grusky2018newsroom}
Grusky, M., Naaman, M., and Artzi, Y. (2018).
\newblock Newsroom: A dataset of 1.3 million summaries with diverse extractive
  strategies.
\newblock {\em arXiv preprint arXiv:1804.11283}.

\bibitem[\protect\citename{Guo et~al., }2018]{Guo:2018:ACL}
Guo, H., Pasunuru, R., and Bansal, M. (2018).
\newblock Soft, layer-specific multi-task summarization with entailment and
  question generation.
\newblock In {\em Proceedings of the Annual Meeting of the Association for
  Computational Linguistics (ACL)}, Melbourne, Australia.

\bibitem[\protect\citename{Harwood, }1996]{Harwood:1996}
Harwood, W.~S. (1996).
\newblock The one minute paper: {A} communication tool for large lecture
  classes.
\newblock {\em Journal of Chemical Education}.

\bibitem[\protect\citename{He et~al., }2012]{He:2012}
He, Z., Chen, C., Bu, J., Wang, C., Zhang, L., Cai, D., and He, X. (2012).
\newblock Document summarization based on data reconstruction.
\newblock In {\em Proceedings of AAAI}.

\bibitem[\protect\citename{Hermann et~al., }2015]{hermann2015teaching}
Hermann, K.~M., Kocisky, T., Grefenstette, E., Espeholt, L., Kay, W., Suleyman,
  M., and Blunsom, P. (2015).
\newblock Teaching machines to read and comprehend.
\newblock In {\em Advances in Neural Information Processing Systems}, pages
  1693--1701.

\bibitem[\protect\citename{Hong et~al., }2014]{HONG14.1093.L14-1070}
Hong, K., Conroy, J., Favre, B., Kulesza, A., Lin, H., and Nenkova, A. (2014).
\newblock A repository of state of the art and competitive baseline summaries
  for generic news summarization.
\newblock In Calzolari, N., Choukri, K., Declerck, T., Loftsson, H., Maegaard,
  B., Mariani, J., Moreno, A., Odijk, J., and Piperidis, S., editors, {\em
  Proceedings of LREC}, pages 1608--1616, Reykjavik, Iceland.
\newblock ACL Anthology Identifier: L14-1070.

\bibitem[\protect\citename{Jindal and Liu, }2008]{jindal2008opinion}
Jindal, N. and Liu, B. (2008).
\newblock Opinion spam and analysis.
\newblock In {\em Proceedings of the 2008 International Conference on Web
  Search and Data Mining}, pages 219--230. ACM.

\bibitem[\protect\citename{Jing and McKeown, }1999]{Jing:1999}
Jing, H. and McKeown, K. (1999).
\newblock The decomposition of human-written summary sentences.
\newblock In {\em Proceedings of the International ACM SIGIR Conference on
  Research and Development in Information Retrieval (SIGIR)}.

\bibitem[\protect\citename{Kikuchi et~al., }2016]{Kikuchi:2016}
Kikuchi, Y., Neubig, G., Sasano, R., Takamura, H., and Okumura, M. (2016).
\newblock Controlling output length in neural encoder-decoders.
\newblock In {\em Proceedings of EMNLP}.

\bibitem[\protect\citename{Lee et~al., }2009]{lee2009automatic}
Lee, J.-H., Park, S., Ahn, C.-M., and Kim, D. (2009).
\newblock Automatic generic document summarization based on non-negative matrix
  factorization.
\newblock {\em Information Processing \& Management}, 45(1):20--34.

\bibitem[\protect\citename{Li et~al., }2013]{Li:2013}
Li, C., Liu, F., Weng, F., and Liu, Y. (2013).
\newblock Document summarization via guided sentence compression.
\newblock In {\em Proceedings of the 2013 Conference on Empirical Methods in
  Natural Language Processing}, pages 490--500, Seattle, Washington, USA.
  Association for Computational Linguistics.

\bibitem[\protect\citename{Li et~al., }2014]{Li:2014:EMNLP}
Li, C., Liu, Y., Liu, F., Zhao, L., and Weng, F. (2014).
\newblock Improving multi-document summarization by sentence compression based
  on expanded constituent parse tree.
\newblock In {\em Proceedings of the Conference on Empirical Methods on Natural
  Language Processing (EMNLP)}, Doha, Qatar.

\bibitem[\protect\citename{Li et~al., }2015]{li-liu-zhao:2015:NAACL-HLT1}
Li, C., Liu, Y., and Zhao, L. (2015).
\newblock Using external resources and joint learning for bigram weighting in
  {ILP}-based multi-document summarization.
\newblock In {\em Proceedings of the 2015 Conference of the North American
  Chapter of the Association for Computational Linguistics: Human Language
  Technologies}, pages 778--787, Denver, Colorado. Association for
  Computational Linguistics.

\bibitem[\protect\citename{Liao et~al., }2018]{Liao:2018}
Liao, K., Lebanoff, L., and Liu, F. (2018).
\newblock Abstract meaning representation for multi-document summarization.
\newblock In {\em Proceedings of the International Conference on Computational
  Linguistics (COLING)}, Santa Fe, New Mexico, USA.

\bibitem[\protect\citename{Lin, }2004]{Lin:2004}
Lin, C.-Y. (2004).
\newblock {ROUGE}: a package for automatic evaluation of summaries.
\newblock In {\em Proceedings of the Workshop on Text Summarization Branches
  Out}, volume~8. Barcelona, Spain.

\bibitem[\protect\citename{Lin and Bilmes, }2010]{Lin:2010:NAACL}
Lin, H. and Bilmes, J. (2010).
\newblock Multi-document summarization via budgeted maximization of submodular
  functions.
\newblock In {\em Human Language Technologies: The 2010 Annual Conference of
  the North American Chapter of the Association for Computational Linguistics},
  pages 912--920. Association for Computational Linguistics.

\bibitem[\protect\citename{Liu and Liu, }2013]{Liu:2013:IEEETrans}
Liu, F. and Liu, Y. (2013).
\newblock Towards abstractive speech summarization: {E}xploring unsupervised
  and supervised approaches for spoken utterance compression.
\newblock {\em IEEE Transactions on Audio, Speech and Language Processing},
  21(7):1469--1480.

\bibitem[\protect\citename{Luo et~al., }2015]{Luo:2015:demo}
Luo, W., Fan, X., Menekse, M., Wang, J., and Litman, D. (2015).
\newblock Enhancing instructor-student and student-student interactions with
  mobile interfaces and summarization.
\newblock In {\em Proceedings of the 2015 Conference of the North American
  Chapter of the Association for Computational Linguistics: Demonstrations},
  pages 16--20, Denver, Colorado. Association for Computational Linguistics.

\bibitem[\protect\citename{Luo and Litman, }2015]{Luo:2015:EMNLP}
Luo, W. and Litman, D. (2015).
\newblock Summarizing student responses to reflection prompts.
\newblock In {\em Proceedings of the 2015 Conference on Empirical Methods in
  Natural Language Processing}, pages 1955--1960, Lisbon, Portugal. Association
  for Computational Linguistics.

\bibitem[\protect\citename{Luo et~al., }2016a]{Luo:2016:COLING}
Luo, W., Liu, F., and Litman, D. (2016a).
\newblock An improved phrase-based approach to annotating and summarizing
  student course responses.
\newblock In {\em Proceedings of COLING 2016, the 26th International Conference
  on Computational Linguistics: Technical Papers}, pages 53--63, Osaka, Japan.
  The COLING 2016 Organizing Committee.

\bibitem[\protect\citename{Luo et~al., }2016b]{Luo:2016:NAACL}
Luo, W., Liu, F., Liu, Z., and Litman, D. (2016b).
\newblock Automatic summarization of student course feedback.
\newblock In {\em Proceedings of the 2016 Conference of the North American
  Chapter of the Association for Computational Linguistics: Human Language
  Technologies}, pages 80--85, San Diego, California. Association for
  Computational Linguistics.

\bibitem[\protect\citename{Martins and Smith, }2009]{martins-smith:2009:ILPNLP}
Martins, A. and Smith, N.~A. (2009).
\newblock Summarization with a joint model for sentence extraction and
  compression.
\newblock In {\em Proceedings of the Workshop on Integer Linear Programming for
  NLP}, pages 1--9, Boulder, Colorado.

\bibitem[\protect\citename{Mazumder et~al., }2010]{Mazumder:2010}
Mazumder, R., Hastie, T., and Tibshirani, R. (2010).
\newblock Spectral regularization algorithms for learning large incomplete
  matrices.
\newblock {\em Journal of Machine Learning Research}.

\bibitem[\protect\citename{Menekse et~al., }2011]{Menekse:2011}
Menekse, M., Stump, G., Krause, S.~J., and Chi, M.~T. (2011).
\newblock The effectiveness of students’ daily reflections on learning in
  engineering context.
\newblock In {\em Proceedings of the American Society for Engineering Education
  (ASEE) Annual Conference}, Vancouver, Canada.

\bibitem[\protect\citename{Moon et~al., }2014]{Moon:2014}
Moon, S., Potdar, S., and Martin, L. (2014).
\newblock Identifying student leaders from mooc discussion forums through
  language influence.
\newblock In {\em Proceedings of EMNLP Workshop on Analysis of Large Scale
  Social Interaction in MOOCs}.

\bibitem[\protect\citename{Mosteller, }1989]{Mosteller:1989}
Mosteller, F. (1989).
\newblock The `muddiest point in the lecture' as a feedback device.
\newblock {\em Teaching and Learning}.

\bibitem[\protect\citename{Nallapati et~al.,
  }2016]{DBLP:journals/corr/NallapatiXZ16}
Nallapati, R., Xiang, B., and Zhou, B. (2016).
\newblock Sequence-to-sequence {RNNs} for text summarization.
\newblock {\em CoRR}, abs/1602.06023.

\bibitem[\protect\citename{Narayan et~al.,
  }2018]{narayan-cohen-lapata:2018:N18-1}
Narayan, S., Cohen, S.~B., and Lapata, M. (2018).
\newblock Ranking sentences for extractive summarization with reinforcement
  learning.
\newblock In {\em Proceedings of the 2018 Conference of the North American
  Chapter of the Association for Computational Linguistics: Human Language
  Technologies, Volume 1 (Long Papers)}, pages 1747--1759, New Orleans,
  Louisiana. Association for Computational Linguistics.

\bibitem[\protect\citename{Nenkova and McKeown, }2011]{Nenkova:2011}
Nenkova, A. and McKeown, K. (2011).
\newblock Automatic summarization.
\newblock {\em Foundations and Trends in Information Retrieval}.

\bibitem[\protect\citename{Paul and James, }2004]{paul2004introduction}
Paul, O. and James, Y. (2004).
\newblock An introduction to {DUC}-2004.
\newblock In {\em Proceedings of the 4th Document Understanding Conference (DUC
  2004)}.

\bibitem[\protect\citename{Paulus et~al., }2017]{Paulus:2017}
Paulus, R., Xiong, C., and Socher, R. (2017).
\newblock A deep reinforced model for abstractive summarization.
\newblock In {\em Proceedings of the Conference on Empirical Methods in Natural
  Language Processing (EMNLP)}.

\bibitem[\protect\citename{Qazvinian et~al., }2013]{Qazvinian:2013}
Qazvinian, V., Radev, D.~R., Mohammad, S.~M., Dorr, B., Zajic, D., Whidby, M.,
  and Moon, T. (2013).
\newblock Generating extractive summaries of scientific paradigms.
\newblock {\em Journal of Artificial Intelligence Research}.

\bibitem[\protect\citename{Qian and Liu, }2013]{qian-liu:2013:EMNLP2}
Qian, X. and Liu, Y. (2013).
\newblock Fast joint compression and summarization via graph cuts.
\newblock In {\em Proceedings of the 2013 Conference on Empirical Methods in
  Natural Language Processing}, pages 1492--1502, Seattle, Washington, USA.
  Association for Computational Linguistics.

\bibitem[\protect\citename{Radev et~al., }2004]{Radev:2004}
Radev, D.~R., Jing, H., Sty{\'{s}}, M., and Tam, D. (2004).
\newblock Centroid-based summarization of multiple documents.
\newblock {\em Inf. Process. Manage.}, 40(6):919--938.

\bibitem[\protect\citename{Rankel, }2016]{rankel2016statistical}
Rankel, P.~A. (2016).
\newblock {\em Statistical analysis of text summarization evaluation}.
\newblock PhD thesis, University of Maryland, College Park.

\bibitem[\protect\citename{Ranzato et~al., }2016]{Ranzato:2016}
Ranzato, M., Chopra, S., Auli, M., and Zaremba, W. (2016).
\newblock Sequence level training with recurrent neural networks.
\newblock In {\em Proceedings of the International Conference on Learning
  Representations (ICLR)}.

\bibitem[\protect\citename{Ren et~al., }2016]{ren-EtAl:2016:COLING}
Ren, P., Wei, F., CHEN, Z., MA, J., and Zhou, M. (2016).
\newblock A redundancy-aware sentence regression framework for extractive
  summarization.
\newblock In {\em Proceedings of COLING 2016, the 26th International Conference
  on Computational Linguistics: Technical Papers}, pages 33--43, Osaka, Japan.
  The COLING 2016 Organizing Committee.

\bibitem[\protect\citename{Rose and Siemens, }2014]{Rose:2014}
Rose, C.~P. and Siemens, G. (2014).
\newblock Shared task on prediction of dropout over time in massively open
  online courses.
\newblock In {\em Proceedings of EMNLP Workshop on Analysis of Large Scale
  Social Interaction in MOOCs}.

\bibitem[\protect\citename{Rus et~al., }2013]{rus:2013}
Rus, V., Lintean, M.~C., Banjade, R., Niraula, N.~B., and Stefanescu, D.
  (2013).
\newblock {SEMILAR}: The semantic similarity toolkit.
\newblock In {\em ACL (Conference System Demonstrations)}, pages 163--168.

\bibitem[\protect\citename{Rush et~al., }2015]{rush-chopra-weston:2015:EMNLP}
Rush, A.~M., Chopra, S., and Weston, J. (2015).
\newblock A neural attention model for abstractive sentence summarization.
\newblock In {\em Proceedings of the 2015 Conference on Empirical Methods in
  Natural Language Processing}, pages 379--389, Lisbon, Portugal. Association
  for Computational Linguistics.

\bibitem[\protect\citename{See et~al., }2017]{see-liu-manning:2017:Long}
See, A., Liu, P.~J., and Manning, C.~D. (2017).
\newblock Get to the point: Summarization with pointer-generator networks.
\newblock In {\em Proceedings of the 55th Annual Meeting of the Association for
  Computational Linguistics (Volume 1: Long Papers)}, pages 1073--1083,
  Vancouver, Canada. Association for Computational Linguistics.

\bibitem[\protect\citename{Song et~al., }2018]{Song:2018}
Song, K., Zhao, L., and Liu, F. (2018).
\newblock Structure-infused copy mechanisms for abstractive summarization.
\newblock In {\em Proceedings of the International Conference on Computational
  Linguistics (COLING)}, Santa Fe, New Mexico, USA.

\bibitem[\protect\citename{Suzuki and Nagata, }2017]{Suzuki:2017}
Suzuki, J. and Nagata, M. (2017).
\newblock Cutting-off redundant repeating generations for neural abstractive
  summarization.
\newblock In {\em Proceedings of the 15th Conference of the European Chapter of
  the Association for Computational Linguistics (EACL)}.

\bibitem[\protect\citename{Takase et~al., }2016]{takase-EtAl:2016:EMNLP2016}
Takase, S., Suzuki, J., Okazaki, N., Hirao, T., and Nagata, M. (2016).
\newblock Neural headline generation on abstract meaning representation.
\newblock In {\em Proceedings of the 2016 Conference on Empirical Methods in
  Natural Language Processing}, pages 1054--1059, Austin, Texas. Association
  for Computational Linguistics.

\bibitem[\protect\citename{Tan et~al., }2017]{Tan:2017}
Tan, J., Wan, X., and Xiao, J. (2017).
\newblock Abstractive document summarization with a graph-based attentional
  neural model.
\newblock In {\em Proceedings of the Annual Meeting of the Association for
  Computational Linguistics (ACL)}.

\bibitem[\protect\citename{Tarnpradab et~al., }2017]{Tarnpradab:2017}
Tarnpradab, S., Liu, F., and Hua, K.~A. (2017).
\newblock Toward extractive summarization of online forum discussions via
  hierarchical attention networks.
\newblock In {\em Proceedings of the 30th Florida Artificial Intelligence
  Research Society Conference (FLAIRS)}, pages 288--292, Marco Island, Florida.

\bibitem[\protect\citename{Taskar, }2012]{Kulesza:2012}
Taskar, A. K.~B. (2012).
\newblock {\em Determinantal Point Processes for Machine Learning}.
\newblock Now Publishers Inc.

\bibitem[\protect\citename{Teufel and Moens, }2002]{Teufel:2002}
Teufel, S. and Moens, M. (2002).
\newblock Summarizing scientific articles: {E}xperiments with relevance and
  rhetorical status.
\newblock {\em Computational Linguistics}.

\bibitem[\protect\citename{Van~den Boom et~al., }2004]{vandenBoom:2004}
Van~den Boom, G., Paas, F., Van~Merrienboer, J.~J., and Van~Gog, T. (2004).
\newblock Reflection prompts and tutor feedback in a web-based learning
  environment: effects on students' self-regulated learning competence.
\newblock {\em Computers in Human Behavior}, 20(4):551 -- 567.

\bibitem[\protect\citename{Vanderwende et~al., }2007]{Vanderwende:2007}
Vanderwende, L., Suzuki, H., Brockett, C., and Nenkova, A. (2007).
\newblock Beyond {S}um{B}asic: Task-focused summarization with sentence
  simplification and lexical expansion.
\newblock {\em Information Processing \& Management}, 43(6):1606--1618.

\bibitem[\protect\citename{Wang et~al., }2008]{wang2008multi}
Wang, D., Li, T., Zhu, S., and Ding, C. (2008).
\newblock Multi-document summarization via sentence-level semantic analysis and
  symmetric matrix factorization.
\newblock In {\em Proceedings of the 31st annual international ACM SIGIR
  conference on Research and development in information retrieval}, pages
  307--314. ACM.

\bibitem[\protect\citename{Wang et~al., }2016a]{wang2016low}
Wang, W.~Y., Mehdad, Y., Radev, D.~R., and Stent, A. (2016a).
\newblock A low-rank approximation approach to learning joint embeddings of
  news stories and images for timeline summarization.
\newblock In {\em Proceedings of the 2016 Conference of the North American
  Chapter of the Association for Computational Linguistics: Human Language
  Technologies}, pages 58--68.

\bibitem[\protect\citename{Wang et~al., }2016b]{wang-EtAl:2016:COLING1}
Wang, X., Nishino, M., Hirao, T., Sudoh, K., and Nagata, M. (2016b).
\newblock Exploring text links for coherent multi-document summarization.
\newblock In {\em Proceedings of COLING 2016, the 26th International Conference
  on Computational Linguistics: Technical Papers}, pages 213--223, Osaka,
  Japan. The COLING 2016 Organizing Committee.

\bibitem[\protect\citename{Wen et~al., }2014a]{Wen:2014:ICWSM}
Wen, M., Yang, D., and Rose, C.~P. (2014a).
\newblock Linguistic reflections of student engagement in massive open online
  courses.
\newblock In {\em Proceedings of ICWSM}.

\bibitem[\protect\citename{Wen et~al., }2014b]{Wen:2014}
Wen, M., Yang, D., and Rosé, C.~P. (2014b).
\newblock Sentiment analysis in {MOOC} discussion forums: {W}hat does it tell
  us?
\newblock In {\em Proceedings of The 7th International Conference on
  Educational Data Mining (EDM)}.

\bibitem[\protect\citename{Wilson, }1986]{Wilson:1986}
Wilson, R.~C. (1986).
\newblock Improving faculty teaching: {E}ffective use of student evaluations
  and consultants.
\newblock {\em Journal of Higher Education}.

\bibitem[\protect\citename{Wolpert, }1996]{wolpert1996lack}
Wolpert, D.~H. (1996).
\newblock The lack of a priori distinctions between learning algorithms.
\newblock {\em Neural computation}, 8(7):1341--1390.

\bibitem[\protect\citename{Xiong, }2015]{xiong2015helpfulness}
Xiong, W. (2015).
\newblock {\em Helpfulness Guided Review Summarization}.
\newblock PhD thesis, University of Pittsburgh.

\bibitem[\protect\citename{Xiong and Litman, }2014]{xiong-litman:2014:Coling}
Xiong, W. and Litman, D. (2014).
\newblock Empirical analysis of exploiting review helpfulness for extractive
  summarization of online reviews.
\newblock In {\em Proceedings of COLING 2014, the 25th International Conference
  on Computational Linguistics: Technical Papers}, pages 1985--1995, Dublin,
  Ireland. Dublin City University and Association for Computational
  Linguistics.

\bibitem[\protect\citename{Yasunaga et~al., }2017]{Yasunaga:2017}
Yasunaga, M., Zhang, R., Meelu, K., Pareek, A., Srinivasan, K., and Radev, D.
  (2017).
\newblock Graph-based neural multi-document summarization.
\newblock In {\em Proceedings of the Conference on Computational Natural
  Language Learning (CoNLL)}, Vancouver, Canada.

\bibitem[\protect\citename{Zhou et~al., }2017]{Zhou:2017}
Zhou, Q., Yang, N., Wei, F., and Zhou, M. (2017).
\newblock Selective encoding for abstractive sentence summarization.
\newblock In {\em Proceedings of the Annual Meeting of the Association for
  Computational Linguistics (ACL)}.

\end{thebibliography}

\appendix
\section{Results without removing low-frequency bigrams}
\label{chapter:ilp_nocutoff}

\begin{table}[!ht]
\begin{center}
\begin{minipage}{0.8\textwidth}
\begin{tabular}{l l| llll }
\hline\hline
 & \multicolumn{1}{l|}{\textbf{System}} & \bf R-1 & \bf R-2  & \bf R-SU4 & \bf R-L\\
\hline
  Eng       &  ILP      &.351           &.108        &.126          &.337\\
            &  ILP+MC   &\bf.355        &\bf.111     &\bf.130       &\bf{.347}  \\\hline
  Stat2015  &  ILP      &.403           &.205        &.163          &.395\\
            &  ILP+MC   &\bf.448$^+$    &\bf.248$^+$ &\bf.214$^+$   &\bf{.433}  \\\hline
  Stat2016  &  ILP      &\bf.470        &\bf.249     &\bf.207       &\bf.455\\
            &  ILP+MC   &.423$^-$       &.209$^-$    &.170$^-$      &.410$^-$  \\\hline
  CS        &  ILP      &.374           &.138        &.139          &.354\\
            &  ILP+MC   &\bf.380        &\bf.144     &\bf.146       &\bf{.362}  \\\hline
  peer      &  ILP      &\bf.470        &\bf.228     &\bf.176       &\bf.449\\
            &  ILP+MC   &.452           &.175        &.165          &.428\\\hline
  camera    &  ILP      &\bf.456        &\bf.168     &\bf.179       &\bf.426\\
            &  ILP+MC   &.440           &.146        &.168          &.410\\\hline
  movie     &  ILP      &.426           &\bf.109     &.163          &.382\\
            &  ILP+MC   &\bf.430        &.102        &\bf.165       &\bf{.397}  \\\hline
  DUC04     &  ILP      &\bf.377        &\bf.092     &\bf.126       &\bf.333\\
            &  ILP+MC   &.337$^-$       &.071$^-$    &.106$^-$      &.305$^-$  \\
\hline\hline
\end{tabular}

\end{minipage}
\end{center}
\caption{Summarization results without removing low-frequency bigrams. That is, all bigrams are used in the matrix approximation process. Compared to Table~\ref{tab:results_ilp}, by using the cutoff technique, both ILP and ILP+MC get better.}
\label{tab:results}
\end{table}

\end{document}